\documentclass[authoryear,review,a4paper]{elsarticle}
\usepackage{amsmath}
\usepackage{mathtools}
\usepackage{adjustbox}
\usepackage{multirow}
\usepackage{textcomp}
\usepackage{lineno,hyperref}
\usepackage{multirow}
\usepackage{tabularx}
\usepackage{caption}
\usepackage{subcaption}
\usepackage{colortbl}
\usepackage{pdflscape}
\usepackage{makecell}
\usepackage{graphicx}
\usepackage{pgfplots}
\usepackage{pdflscape}
\usepackage{longtable}
\usepackage{pgfplotstable}

\pgfplotstableset{
    /color cells/min/.initial=0,
    /color cells/max/.initial=1000,
    /color cells/textcolor/.initial=,
    color cells/.code={%
        \pgfqkeys{/color cells}{#1}%
        \pgfkeysalso{%
            postproc cell content/.code={%
                \begingroup
                \pgfkeysgetvalue{/pgfplots/table/@preprocessed cell content}\value
\ifx\value\empty
\endgroup
\else
                \pgfmathfloatparsenumber{\value}%
                \pgfmathfloattofixed{\pgfmathresult}%
                \let\value=\pgfmathresult
                \pgfplotscolormapaccess
                    [\pgfkeysvalueof{/color cells/min}:\pgfkeysvalueof{/color cells/max}]%
                    {\value}%
                    {\pgfkeysvalueof{/pgfplots/colormap name}}%
                \pgfkeysgetvalue{/pgfplots/table/@cell content}\typesetvalue
                \pgfkeysgetvalue{/color cells/textcolor}\textcolorvalue
                \toks0=\expandafter{\typesetvalue}%
                \xdef\temp{%
                    \noexpand\pgfkeysalso{%
                        @cell content={%
                            \noexpand\cellcolor[rgb]{\pgfmathresult}%
                            \noexpand\definecolor{mapped color}{rgb}{\pgfmathresult}%
                            \ifx\textcolorvalue\empty
                            \else
                                \noexpand\color{\textcolorvalue}%
                            \fi
                            \the\toks0 %
                        }%
                    }%
                }%
                \endgroup
                \temp
\fi
            }%
        }%
    }
}

\modulolinenumbers[5]
\begin{document}

\begin{frontmatter}

\title{An overview of deep learning techniques for epileptic seizures detection and prediction based on neuroimaging modalities: Methods, challenges, and future works}

\author[graaf]{Afshin Shoeibi \corref{mycorrespondingauthor}}
\cortext[mycorrespondingauthor]{Corresponding author}
\ead{afshin.shoeibi@gmail.com}
\author[mor]{Parisa Moridian}
\ead{parisamoridian@yahoo.com}
\author[mhd]{Marjane Khodatars}
\ead{khodatars1marjane@gmail.com}
\author[ferdowsi]{Navid Ghassemi}
\ead{navidghassemi@mail.um.ac.ir}
\author[jaf]{Mahboobeh Jafari}
\ead{mahbube.jafari@yahoo.com}
\author[rol]{Roohallah Alizadehsani}
\ead{ralizadehsani@deakin.edu.au}
\author[kon]{Yinan Kong}
\ead{yinan.kong@mq.edu.au}
\author[graaf]{Juan Manuel Gorriz}
\ead{gorriz@ugr.es}
\author[rem]{Javier Ramírez}
\ead{javierrp@ugr.es}
\author[rol]{Abbas Khosravi}
\ead{abbas.khosravi@deakin.edu.au}
\author[rol]{Saeid Nahavandi}
\ead{saeid.nahavandi@deakin.edu.au}
\author[achone,achtwo,achthree]{U. Rajendra Acharya}
\ead{Rajendra_Udyavara_ACHARYA@np.edu.sg}

\address[graaf]{Data Science and Computational Intelligence Institute, University of Granada, Spain.}
\address[mor]{Faculty of Engineering, Science and Research Branch, Islamic Azad University, Tehran, Iran.}
\address[mhd]{Department of Medical Engineering, Mashhad Branch, Islamic Azad University, Mashhad, Iran.}
\address[ferdowsi]{Computer Engineering department, Ferdowsi University of Mashhad, Mashhad, Iran.}
\address[jaf]{Electrical and Computer Engineering Faculty, Semnan University, Semnan, Iran.}
\address[rol]{Intelligent for Systems Research and Innovation (IISRI), Deakin University, Victoria 3217, Australia.}
\address[kon]{School of Engineering, Macquarie University, Sydney 2109, Australia.}
\address[rem]{Department of Signal Theory, Networking and Communications, Universidad de Granada, Spain.}
\address[achone]{Ngee Ann Polytechnic, Singapore 599489, Singapore.}
\address[achtwo]{Dept. of Biomedical Informatics and Medical Engineering, Asia University, Taichung, Taiwan.}
\address[achthree]{Dept. of Biomedical Engineering, School of Science and Technology, Singapore University of Social Sciences, Singapore.}

\begin{abstract}

Epilepsy is a disorder of the brain denoted by frequent seizures. The symptoms of seizure include confusion, abnormal staring, and rapid, sudden, and uncontrollable hand movements. Epileptic seizure detection methods involve neurological exams, blood tests, neuropsychological tests, and neuroimaging modalities. Among these, neuroimaging modalities have received considerable attention from specialist physicians. One method to facilitate the accurate and fast diagnosis of epileptic seizures is to employ computer-aided diagnosis systems (CADS) based on deep learning (DL) and neuroimaging modalities. This paper has studied a comprehensive overview of DL methods employed for epileptic seizures detection and prediction using neuroimaging modalities. First, DL-based CADS for epileptic seizures detection and prediction using neuroimaging modalities are discussed. Also, descriptions of various datasets, preprocessing algorithms, and DL models which have been used for epileptic seizures detection and prediction have been included. Then, research on rehabilitation tools has been presented, which contains brain-computer interface (BCI), cloud computing, internet of things (IoT), hardware implementation of DL techniques on field-programmable gate array (FPGA), etc. In the discussion section, a comparison has been carried out between research on epileptic seizure detection and prediction. The challenges in epileptic seizures detection and prediction using neuroimaging modalities and DL models have been described. In addition, possible directions for future works in this field, specifically for solving challenges in datasets, DL, rehabilitation, and hardware models, have been proposed. The final section is dedicated to the conclusion which summarizes the significant findings of the paper.

\end{abstract}      


\end{frontmatter}

\section{Introduction}

\begin{figure}[h]
    \centering
    \includegraphics[width=\textwidth ]{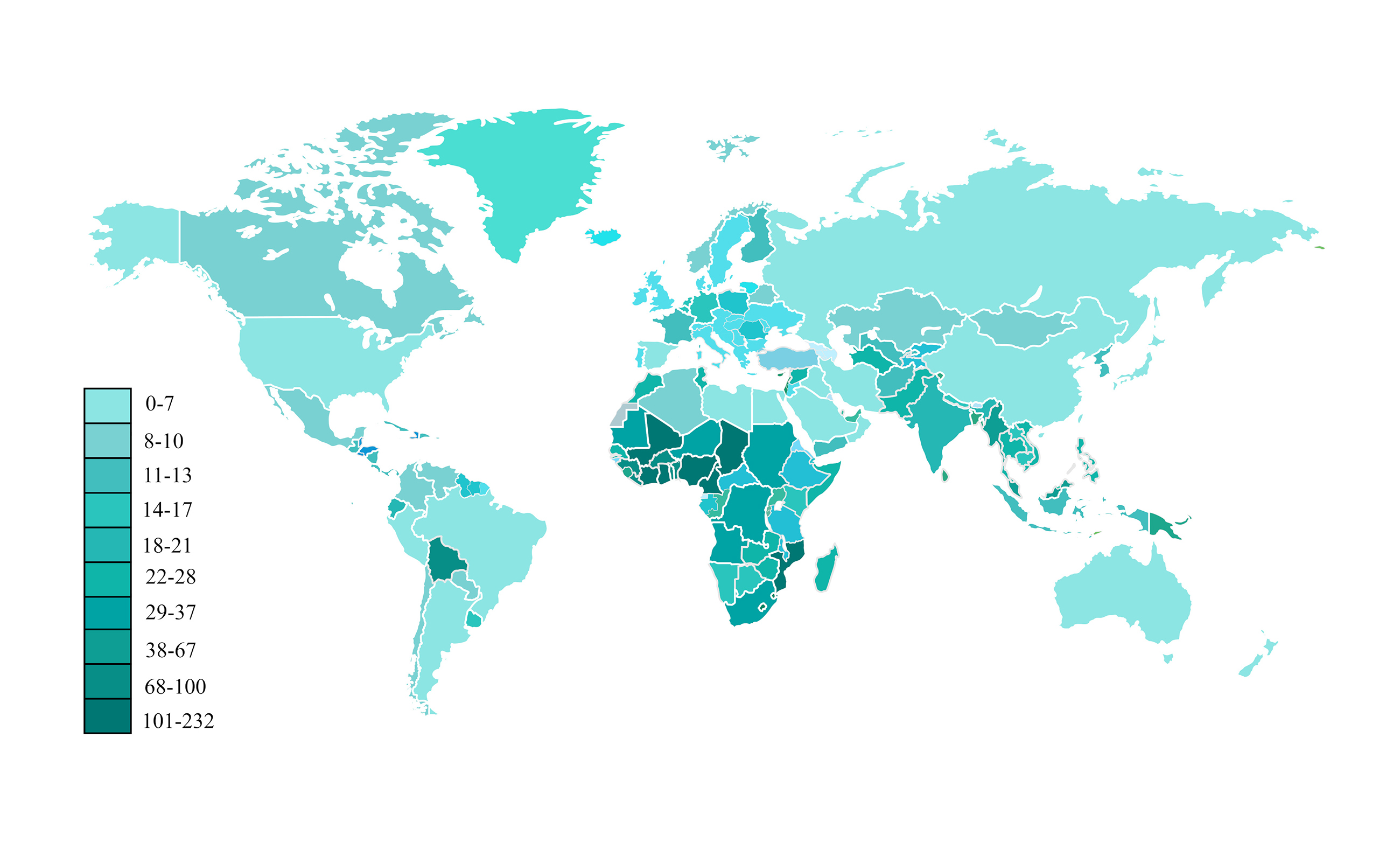}
    
    \caption{Graph of the number of people with epileptic seizures worldwide \citep{a9}.}
    \label{fig:one}  
\end{figure}

Epileptic seizures are a non-communicable disease and are one of the most prevalent disorders of the nervous system. Epileptic disorders usually occur with sudden attacks that result from abnormal activity of the cortical or membrane nerve in the brain \citep{a1,a2,a3,a4,a5}. More than 60 million people worldwide have various types of epileptic seizures and suffer from them \citep{a6,a7,a8}. Figure \ref{fig:one} displays the number of people with epileptic seizures in various parts of the world \citep{a9}. As shown in the figure, the number of people with this type of neurological disorder is greater in underdeveloped countries than in other countries \citep{a9}.

Epilepsy is a rapid and early abnormality in the brain's electrical activity, disrupting part or all of the human body \citep{a10,a11}. Medical researchers have divided epileptic seizures into three categories: generalized \citep{a12,a13}, focal \citep{a14,a15}, and epilepsy with unknown onset \citep{a16}, each of which has various types. 

\begin{figure}[h]
    \centering
    \includegraphics[width=\textwidth ]{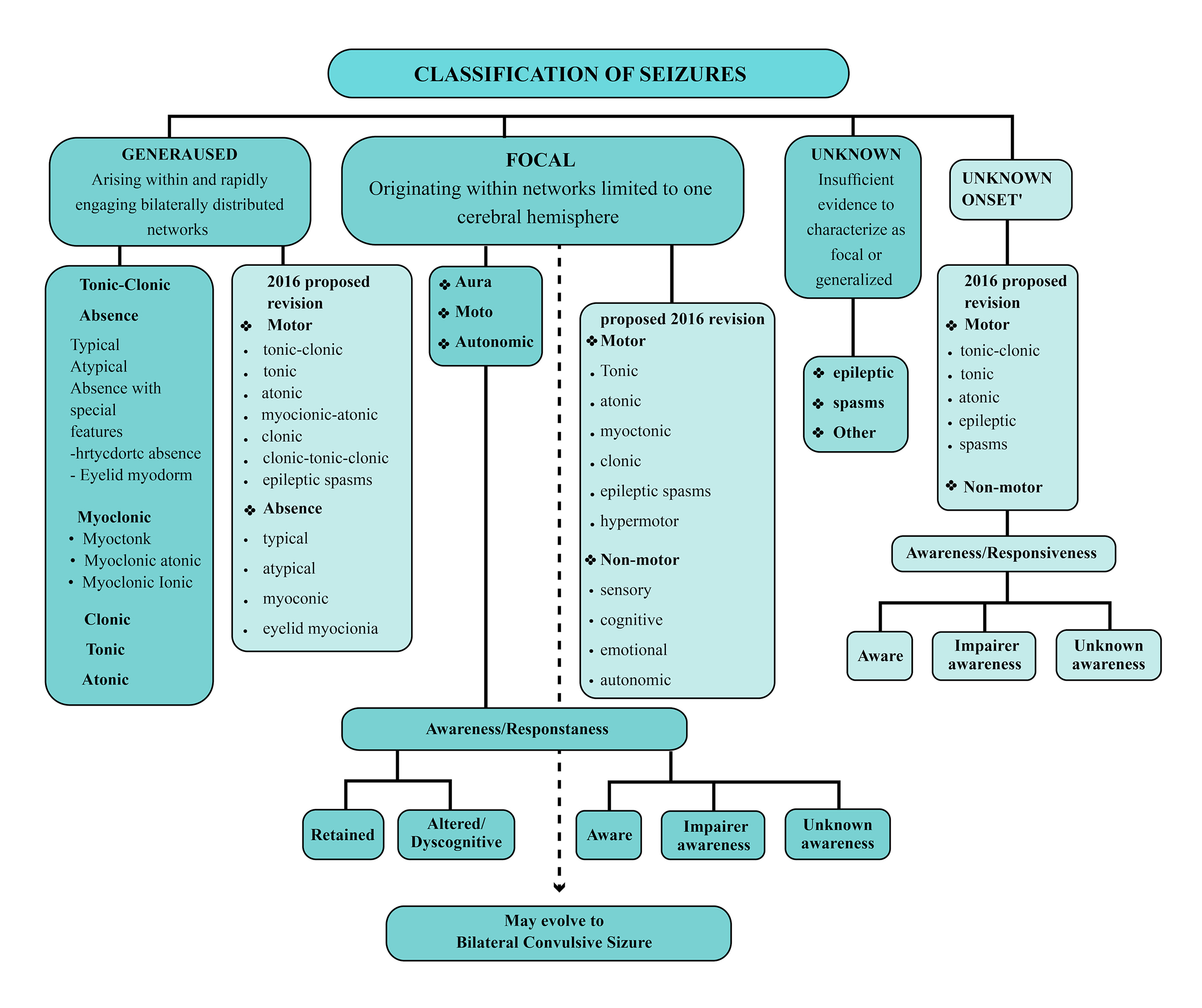}
    
    \caption{Showing different types of general and focal epileptic seizures \citep{a16}.}
    \label{fig:two}
\end{figure}

General epilepsy involves the whole brain and causes disruption of the activity of all neurons in the brain, eventually may lead to the impairment of all parts of the brain \citep{a17,a18,a19}. In partial epilepsy, a small group of neurons form a focal epilepsy and are confined to a hemisphere of the brain. 60\% of patients with epilepsy have focal seizures that are mostly drug-resistant \citep{a20,a21,a22}. The classification of epileptic seizure types is shown in Figure \ref{fig:two}. In this figure, the classification of epileptic seizures before 2011 is depicted in darker color, and the classification of epileptic seizures from 2016 onwards is highlighted in lighter color. More information is available in reference \citep{a16}. 

People with epileptic seizures may sometimes experience severe psychological trauma due to embarrassment and lack of proper social status \citep{a23,a24}. Given the above, accurate and rapid diagnosis of these neurological disorders in the early stages is crucially pivotal.

\begin{figure}[h]
\centering
\begin{subfigure}[b]{0.45\textwidth}
    \centering
    \includegraphics[width=\textwidth]{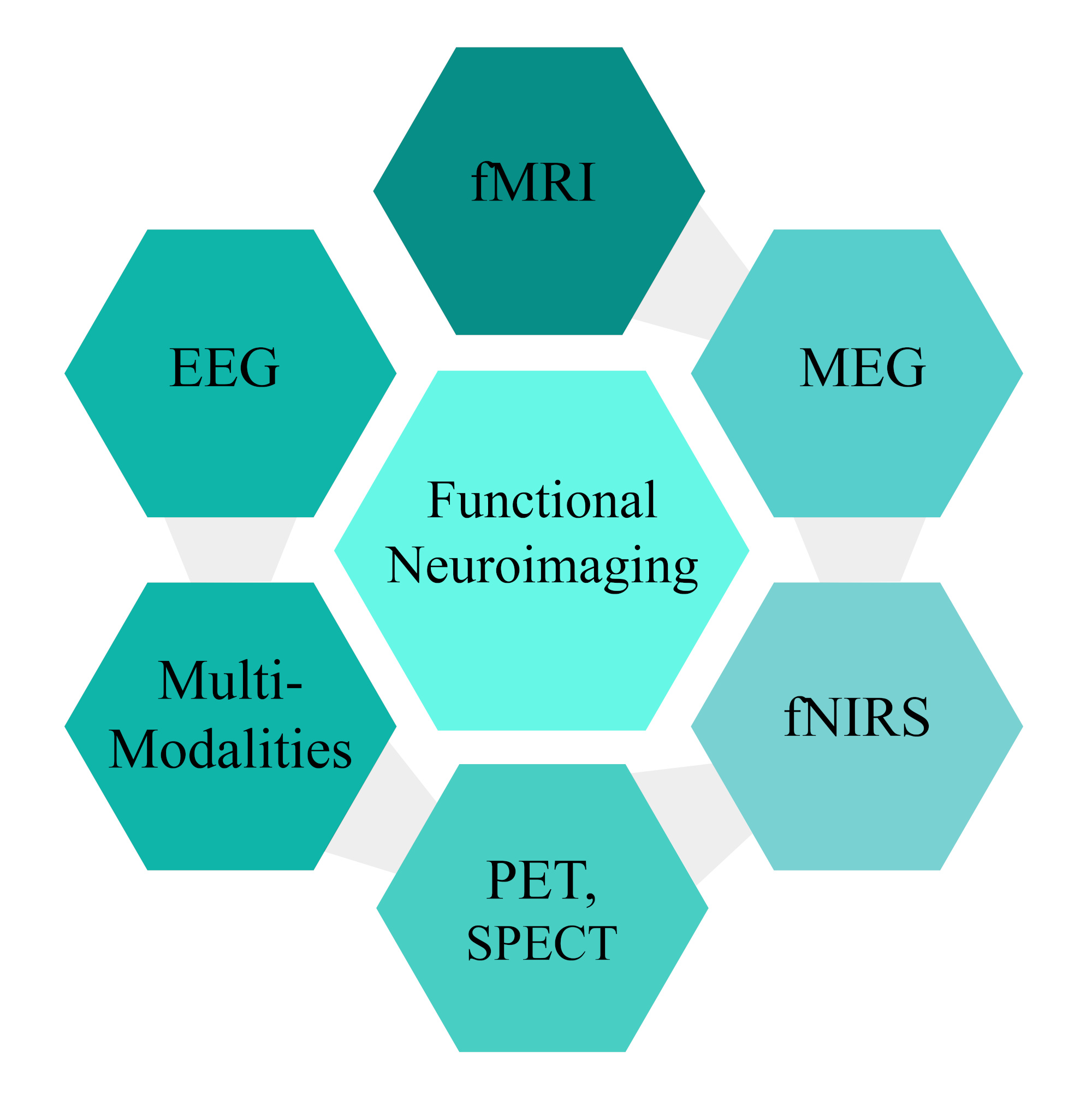}
    \caption{}
    \label{fig:3a}
\end{subfigure}
\hspace{10pt}
\begin{subfigure}[b]{0.45\textwidth}
    \centering
    \includegraphics[width=\textwidth]{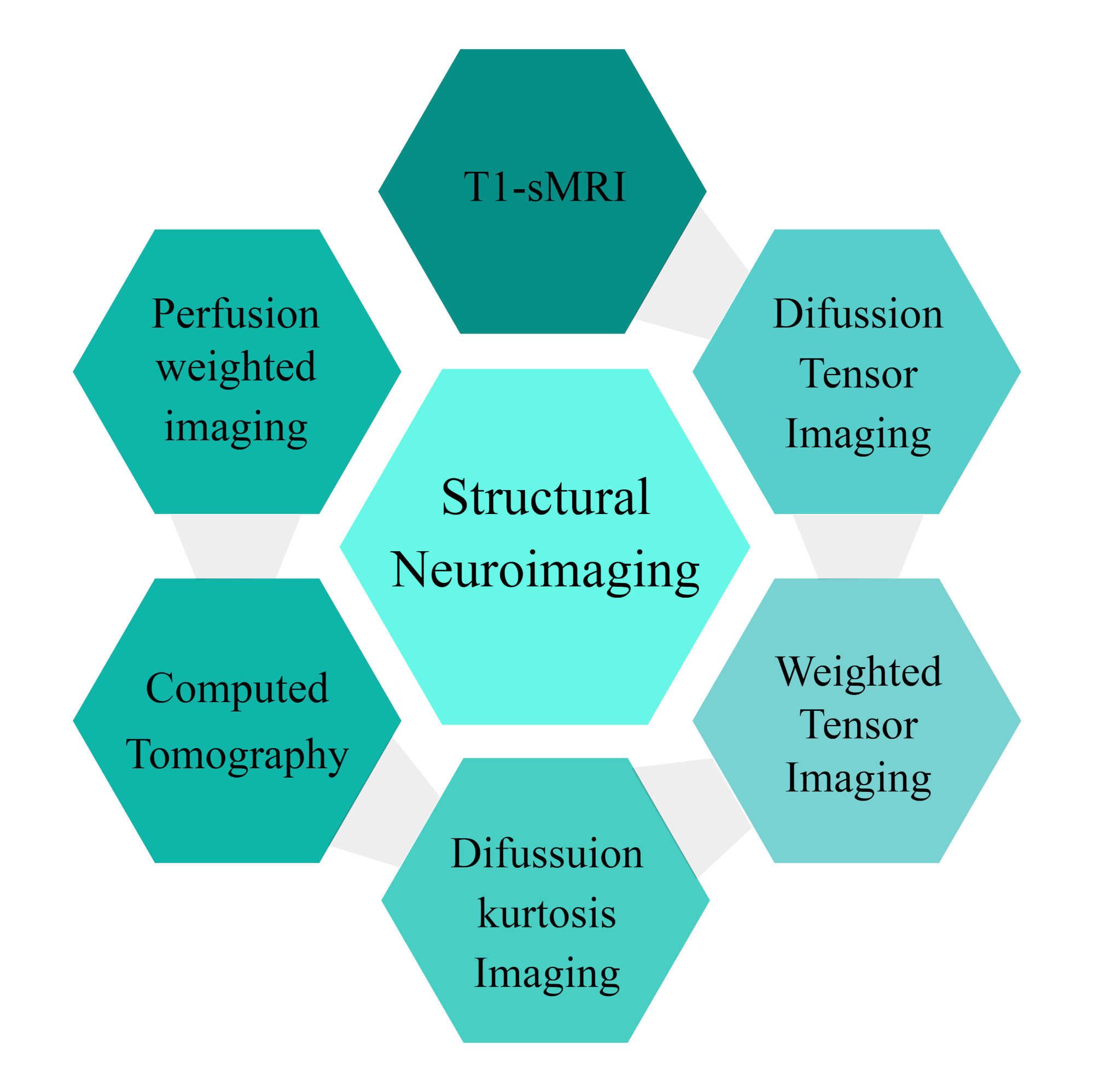}
    \caption{}
    \label{fig:3b}
\end{subfigure}
\caption{Neuroimaging modalities for epileptic seizures detection.}
\label{fig:3}
\end{figure}

Specialist physicians usually use functional and structural neuroimaging techniques to diagnose epileptic seizures. Electroencephalogram (EEG) \citep{a25,a26}, functional magnetic resonance imaging (fMRI) \citep{a27,a28}, magnetoencephalography (MEG) \citep{a29,a30}, electrocorticography (ECoG) \citep{a31,a32}, functional near-infrared spectroscopy (fNIRS) \citep{a33,a34}, positron emission tomography (PET) \citep{a35,a36}, and SPECT \citep{a37} are the most substantial functional neuroimaging modalities. In contrast, structural MRI (sMRI) \citep{a38,a39} and diffusion tensor imaging (DTI) \citep{a40,a41} are in the category of structural neuroimaging modalities. In figure \ref{fig:3}, neuroimaging modalities for epileptic seizures detection are described.

As shown in Figure \ref{fig:3}, structural neuroimaging modalities include sMRI and DTI approaches. By using the sMRI modality, structural abnormalities and brain lesions caused by epileptic seizures can be identified \citep{a42,a43}. Additionally, this modality can be used to identify the anatomical zone of the epileptogenic region that is responsible for the seizure, which is a pivotally significant step for presurgical evaluations of epilepsy \citep{a42,a43}. sMRI is also employed after surgery to evaluate the success or failure of the epileptic region removal and to assess the need for reoperation \citep{a42,a43}. Disadvantages of sMRI include its widespread unavailability, high cost, and the necessity for long-term scans.

The DTI modality provides information on the structural anatomy of the white matter tracts and makes it possible to investigate the microstructural status of the white matter. Although the advantages of applying DTI in the diagnosis of the lesions for epilepsy are still being examined, modeling and reconstructing hidden pathways in white matter is of utmost importance as a presurgical evaluation step \citep{a44}.

In EEG modality, the measurement of voltage fluctuations produced by the ionic current of neurons in the brain is performed, indicating the bioelectric activity of the brain and containing the physiological information of people with epileptic seizures \citep{a45,a46}. The investigations reviewed in this paper demonstrates the effectiveness of EEG modality performance in diagnosing epileptic seizures. EEG incorporates two methods of non-invasive scalp (sEEG) \citep{a47,a48} and intracranial (IEEG) recording \citep{a49,a50}. The sEEG method is widely used by specialist physicians and neurologists compared to IEEG due to its lower risks and more straightforward recording. Additionally, considering that these signal recordings are economically inexpensive and the fact that the frequency and rhythm of brain activity vary during seizures, EEG has become one of the foremost epileptic seizures diagnostic methods \citep{a51,a52}. Compared to EEG, ECoG, fNIRS, and MEG functional modalities are less effective in diagnosing epileptic seizures.

fMRI modality is another neuroimaging technique for epileptic seizures detection and includes two methods based on task (T-fMRI) \citep{a53} and resting state (rs-fMRI) \citep{a54}. fMRI is adapted to detect changes in regional blood flow and metabolism due to the activation of brain regions \citep{a53,a54}.  One of the fMRI applications in epilepsy is identifying ictal and interictal phenomena given rise to the localization of the focal seizures \citep{a53,a54}. During seizures, brain function changes in the epileptogenic region, which can be detected using fMRI \citep{a53,a54}. fMRI can also be exploited to assess brain function before surgery in patients with drug-resistant epilepsy \citep{a53,a54}. One of the drawbacks of fMRI is that the patient has to be in the scanner for a long period to seizure occur and the scan to be completed.

Detecting epileptic seizures from neuroimaging modalities with all the benefits that are sometimes challenging. Epileptic seizure detection using neuroimaging modalities requires a considerable amount of recording data in order for the specialist doctors to make the appropriate decisions. Big data analysis of neuroimaging modalities in most cases beget incorrect epileptic seizures diagnosis by physicians. This is due to eye fatigue when interpreting many structural or functional imaging modalities. Additionally, the presence of diverse noises in neuroimaging modalities is another cause of misdiagnosis. In order to conquer these dilemmas, CADS for epileptic seizures detection using neuroimaging modalities and AI are of considerable help to specialists in the epileptic seizures detection field.

So far, many research works have been conducted to diagnose epileptic seizures using AI. Until quite a few years ago, most examinations were performed in the field of conventional machine learning \citep{a55,a56}. In traditional machine learning, the selection of the feature extraction, reduction and classification techniques is dependent on the characteristics of the data \citep{a57,a58}. However, in DL approaches, all these steps are fulfilled via integrated layers and automatically \citep{a59,a60}. Various DL methods have promptly received a tremendous amount of attention from numerous experts in the brain signal processing domain \citep{a61}. This has made the diagnosis of epileptic seizures based on functional and structural brain modalities along with DL techniques one of the most novel areas of research. In this paper, a complete review of conducted research in the epileptic seizures field from neuroimaging modalities along with DL methods, along with challenges and future work in this field has been presented.

\begin{figure}[h]
    \centering
    \includegraphics[width=\textwidth ]{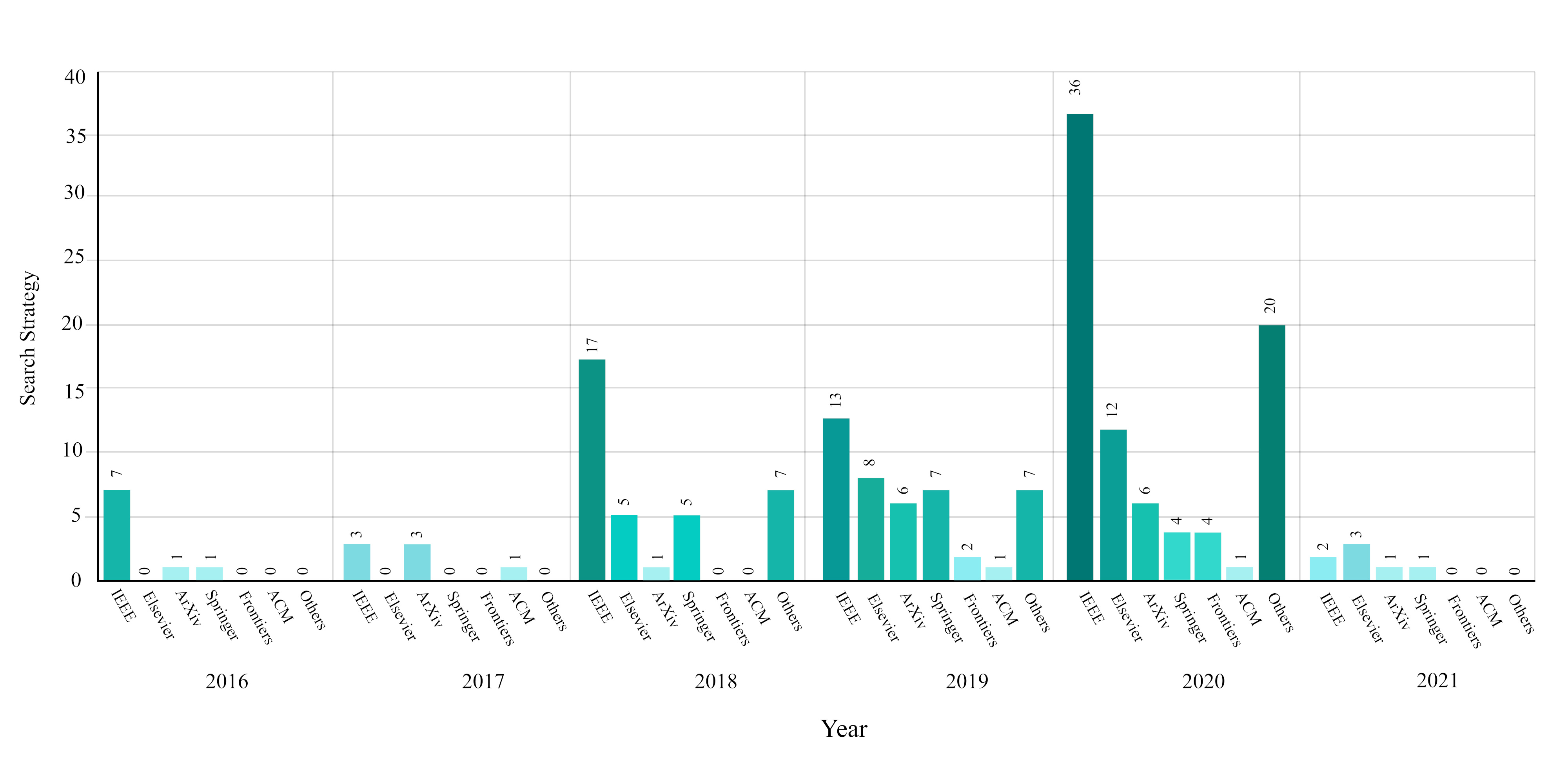}
    
    \caption{Number of papers published for automated detection of epileptic seizures using DL techniques.}
    \label{fig:4}
\end{figure}

In order to search for papers in the scope of diagnosis of epileptic seizures, various citation databases such as IEEE Xplore, ScienceDirect, SpringerLink, and Wiley have been exploited. In addition, Google Scholar has been used to find papers with the keywords "Epileptic Seizure," "EEG," "fMRI," "ECoG', "MEG," "fNIRS," "MRI," "PET" and "Deep Learning." The latest articles were reviewed by the authors on December 30th, 2020. The number of papers accepted each year in different citation sites for the diagnosis of epileptic seizures is illustrated in Figure \ref{fig:4}.

In the following, the outline of this investigation is introduced. The second section concisely presents the DL networks exploited in diagnosing epileptic seizures. Recent CADS for epileptic seizures using DL techniques are examined in Section three. Several research works in the field of rehabilitation systems, cloud computing, and diagnostic epilepsy procedures using non-neural modalities are presented in the fourth section. The discussion is introduced in Section five. In the sixth section, the challenges in diagnosing epileptic seizures are fully described. Finally, conclusions and recommendations for future work are provided in the seventh section.



\section{Epileptic Seizures Detection Using DL Techniques }

\begin{figure}[h]
    \centering
    \includegraphics[width=\textwidth ]{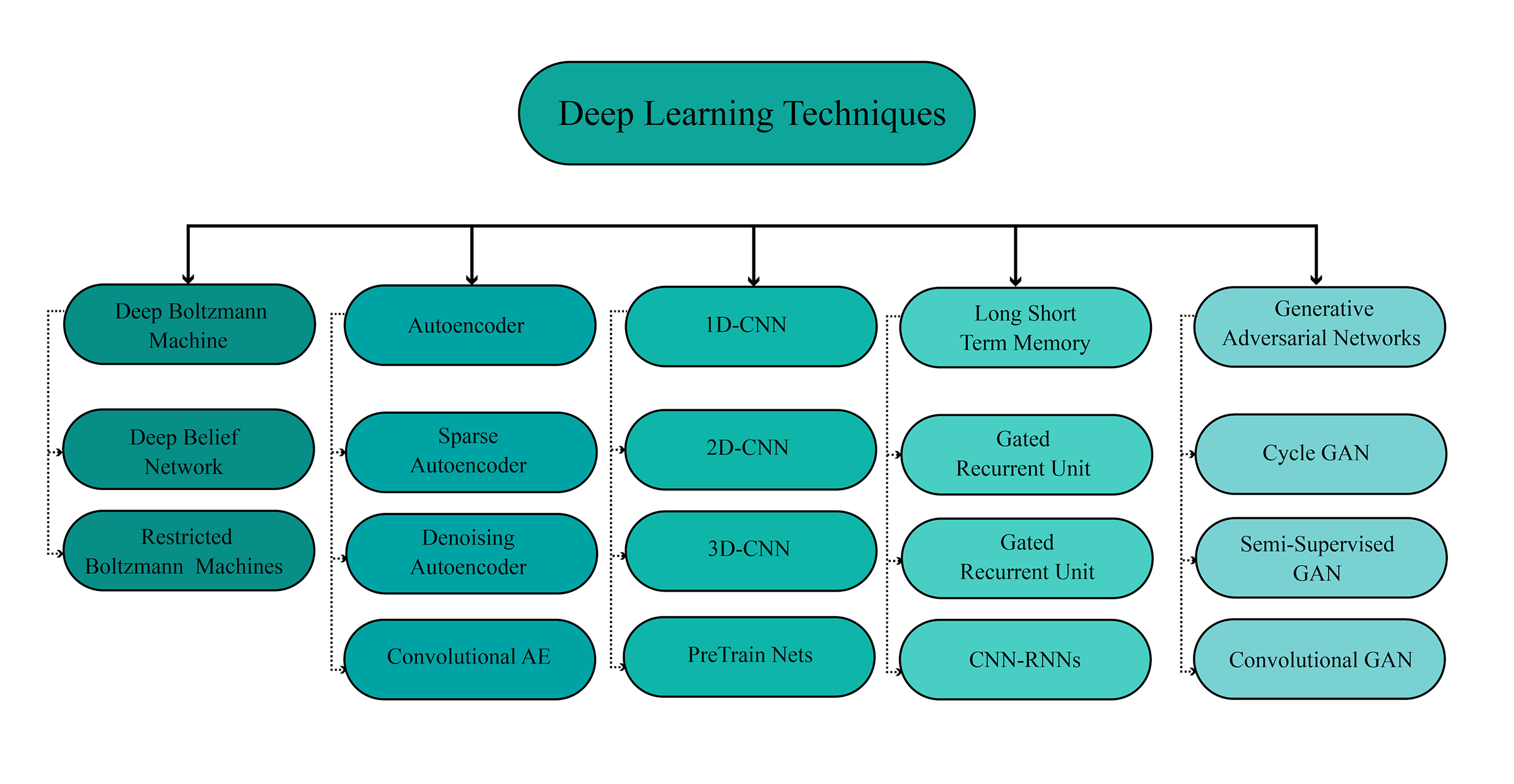}
    
    \caption{Illustration of various DL methods used for epileptic seizures detection.}
    \label{fig:5}
\end{figure}

In this section, DL networks used in the diagnosis of epileptic seizures are presented. Convolutional neural networks (CNNs) are the first category of DL architectures involving a variety of one-dimensional (1D), two-dimensional (2D), and three-dimensional (3D) models \citep{a60,n5}. These networks follow supervised learning and have three main layers: convolutional, pooling, and fully connected (FC) layers \citep{a60}. Recurrent neural networks (RNNs) are another paradigm of DL networks that are based on supervised learning widely applied in time series tasks \citep{a60,n6}. Autoencoders (AEs) models \citep{a60,a62} and deep belief networks (DBNs) \citep{a60,a63} are other types of DL networks based on unsupervised learning. In addition to these models, improved methods from CNN named generative adversarial networks (GANs) \citep{a64} architectures have been proposed for various applications that are based on unsupervised learning. It should be pointed out that generative adversarial networks (GANs) architectures are adopted as supervised techniques in some issues \citep{a64,a65,n7,n2}. CNN-RNN and CNN-AE are two other categories of DL systems created by combining two different architectures \citep{a66}. The CNN-RNN and CNN-AE architectures follow supervised and unsupervised learning, respectively \citep{a67}. Details of the types of DL networks in the diagnosis of epileptic seizures are manifested in Figure \ref{fig:5}. 

\section{CAD Based on DL techniques for Epileptic Seizures using Neuroimaging Modalities}

\begin{figure}[h]
    \centering
    \includegraphics[width=\textwidth ]{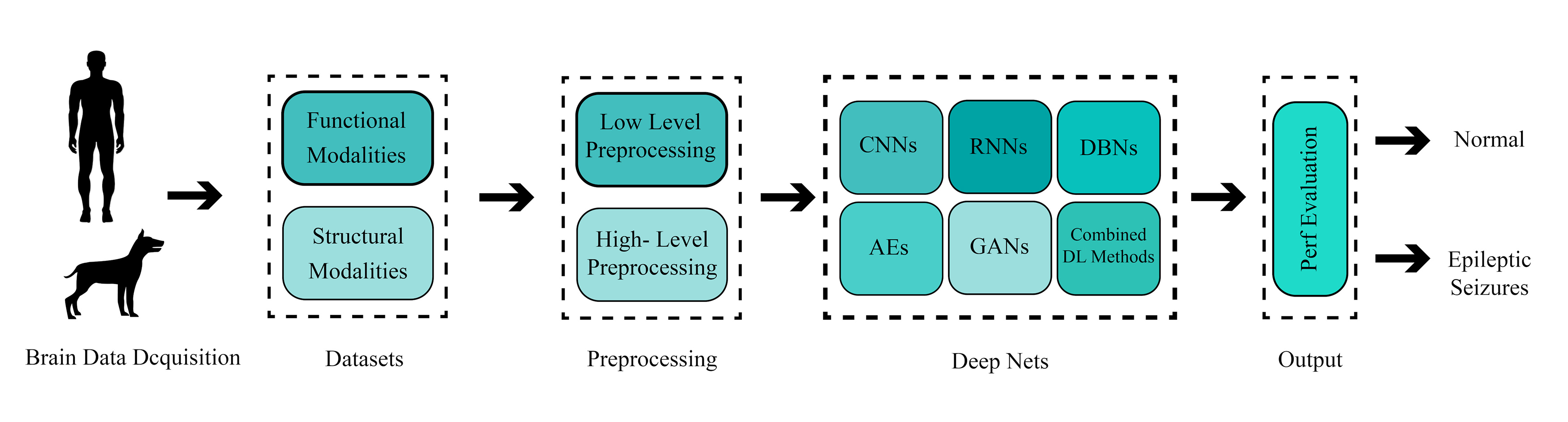}
    
    \caption{Illustration of block diagram for epileptic seizures detection using DL methods.}
    \label{fig:6}
\end{figure}

Diagnosis of epileptic seizures from functional and structural neuroimaging modalities of the brain along with AI algorithms has a long history. Until recently, the diagnosis of epileptic seizures using CADS was based on conventional machine learning techniques that have been the subject of much research \citep{a68,a69,a70,a71}. The most significant weaknesses of these systems were the process of selecting the best feature extraction and dimensional reduction algorithms (feature selection or reduction) using trial and error that required a considerable amount of knowledge in the AI fields \citep{n1}. To resolve these issues, from 2016 onwards, DL methods in CADS for epileptic seizures detection were considered and quickly replaced the conventional machine learning approaches. In CADS based-DL, the feature extraction and selection steps are accomplished entirely automatically. The CADS for epileptic seizures detection based on DL techniques are represented in figure \ref{fig:6}. 

According to figure \ref{fig:6}, a variety of structural and functional neuroimaging modalities are first considered as DL input. In the following, low-level and high-level preprocessing methods are applied to the input data. Then, feature extraction up to classification steps are performed by the desired DL networks (DL networks for epileptic seizures detection research papers are displayed in Appendix A). Finally, various evaluation parameters such as accuracy, sensitivity, and precision are calculated.

\subsection{Epileptic Seizures Datasets}

In this section, the most notable datasets on diagnosing epileptic seizures are reviewed, all of which are freely accessible. Without proper datasets, developing accurate and robust CADS is not possible. Several EEG datasets and an ECoG dataset are currently available to researchers freely; however, datasets on other neuroimaging modalities such as MRI have not yet been made freely available. Multiple EEG datasets, namely Freiburg \citep{a72}, CHB-MIT \citep{a73}, Kaggle \citep{a174}, Bonn \citep{a75}, Flint-Hills \citep{a76}, Bern-Barcelona \citep{a77}, Hauz Khas \citep{a76}, and Zenodo \citep{a78} are the main ones for developing automatic systems for epileptic seizure detection. The signals forming each datapoint of these datasets are recorded either intracranial or from the scalp of humans or animals. Table \ref{tab:data} provides the supplementary information on each dataset, and also, the types of EEG datasets for epileptic seizures diagnosis using DL are listed in table \ref{tab:related}.

\begin{table}[!h]
    \centering
    \caption{List of popular epileptic seizure datasets.}
    \label{tab:data}
    \resizebox{\textwidth}{!}{
    \begin{tabular}{|c|c|c|c|c|c|}
    \hline
    \textbf{Dataset} & \makecell{\textbf{Number of}\\\textbf{Patient}} & \makecell{\textbf{Number of}\\\textbf{Seizures}} & \textbf{Recording} & \makecell{\textbf{Total}\\ \textbf{Duration}\\\textbf{(hour)}} & \makecell{\textbf{Sampling}\\\textbf{Frequency}\\\textbf{(Hz)}}\\
    \hline
\makecell{Flint-Hills\\ \citep{a76}} & 10 & 59 & Continues intracranial ling term ECoG & 1419 & 249 \\ \hline
\makecell{Hauz Khas\\ \citep{a76}} & 10 & NA & Scalp EEG (sEEG) & NA & 200 \\ \hline
\makecell{Freiburg\\ \citep{a72}} & 21 & 87 & Intracranial Electroencephalography (IEEG) & 708  & 256 \\ \hline
\makecell{CHB-MIT\\ \citep{a73}} & 22 & 163  & sEEG & 844  & 256    \\ \hline
\makecell{Kaggle\\ \citep{a174}} & \makecell{5 Dogs\\ \hline 2 Patients} & 48 & IEEG & 622  & \makecell{400\\ \hline 5000}\\ \hline
\makecell{Bonn\\ \citep{a75}} & 10 & NA & Surface and IEEG & 39m  & 173.61 \\ \hline
\makecell{Bern Barcelona\\ \citep{a77}}& 5 & 3750 & IEEG & 83 & 512 \\ \hline
\makecell{Zenodo\\ \citep{a78}} & 79 Neonatal & 460  & sEEG & 74m  & 256 \\

    \hline
    \end{tabular}}
    
\end{table}
\subsection{Preprocessing}

\subsubsection{EEG Preprocessing}

Preprocessing is the first step in DL-based CADS for epileptic seizures detection. The presence of different artifacts in EEG signals always poses a severe challenge to physicians and neurologists in accurately diagnosing epileptic seizures. Artifacts from eye blinks, eye movements, muscle expansion and contraction, and municipal electricity noise are among the most important EEG data noises that should be eliminated from the signals in the preprocessing step \citep{a79,a80,a81,a82}. In some cases, the presence of multiple artifacts begets loss of EEG signals' substantial information between various noises and makes it challenging to diagnose epileptic seizures. EEG signal preprocessing in the diagnosis of epileptic seizures is divided into two types of low-level and high-level approaches, which are explained following. Table \ref{tab:related} shows the low-level and high-level preprocessing techniques of EEG signals in epileptic diagnostic research.

\textbf{A. Low Level EEG Preprocessing}

In this section, low-level preprocessing methods are presented in the DL-based CADS for epileptic seizure detection. Low-level preprocessing in EEG signals involves noise removal, normalization, down-sampling, and segmentation. In order to remove noise from EEG signals, various types of low-pass, high-pass, and band-pass based Butterworth and or Chebyshev filters with different orders are widely employed (these filters are of finite impulse response (FIR) or Infinite impulse response (IIR) types) \citep{a83,a84,a85}. Raw EEG signals have variable voltage amplitude degrading the efficiency of CADS in diagnosing epileptic seizures. To obviate this problem, it is recommended to utilize different normalization methods such as Z-Score \citep{a86}. Storing and processing EEG signals requires a lot of memory space. By using down-sampling, EEG signals sampling frequency is decreased by half, which halves the storage space of EEG signals. Windowing or segmentation of EEG signals is the last part of low-level preprocessing. Segmentation assists in decomposing EEG data into more detailed sections to extract more significant information from each signal frame \citep{a87}.

\textbf{B. High Level EEG Preprocessing}

High-level preprocessing techniques play a pivotal role in enhancing the efficiency of CADS in diagnosing epileptic seizures. In this section, Data augmentation (DA) models are stated as the first category of high-level preprocessing \citep{a88,a89}. The deficiency of EEG signals usually causes overfitting of DL networks during training, and the exploiting of DA techniques is a proper approach to address this problem. Discrete wavelet transform (DWT) \citep{a90}, continues wavelet transform (CWT) \citep{a91}, fast Fourier transform (FFT) \citep{a92}, and empirical mode decomposition (EMD) \citep{a93} are other high-level preprocessing techniques employed to eliminate noise and extract meaningful frequency bands from EEG signals. In addition, some improved FFT techniques such as short-time Fourier transform (STFT) to transform EEG signals to 2D images for application to CNNs have been investigated in research \citep{a94}. Also, some studies have selected independent component analysis (ICA)-based techniques to preprocess the EEG signals of epileptic seizures and have achieved satisfactory results \citep{a95}. Feature extraction procedures have also been considered in research as a crucial step in high-level preprocessing \citep{a293}.

\textbf{C. Medical Imaging Modalities Preprocessing}

Medical imaging modalities are another method of diagnosing epileptic seizures that possesses a special place among specialist physicians. In imaging techniques, applying preprocessing techniques is of great significance. According to Table \ref{tab:rel2}, epileptic seizure detection using MRI modalities is more significant than other techniques. MRI neuroimaging modalities contain structural (sMRI) and functional (fMRI) techniques \citep{a96}. In sMRI modalities, the most important low-level preprocessing techniques include denoising, inhomogeneity correction, brain extraction, registration, intensity standardization, and re-orientation \citep{a97,a98}. Also, slice timing correction, motion correction, normalization, smoothing, and filtering are the most important low-level fMRI preprocessing methods \citep{a99,a100}. Some of the high-level preprocessing methods that have been surveyed in investigations for sMRI and fMRI modalities are segmentation \citep{a101} and functional connectivity matrix (FCM) \citep{a102}, respectively. Other research has focused on using PET imaging modality for diagnosis \citep{a301,a302}. ROI, normalization, Ordered subset expectation maximization (OSEM), and down-sampling are some of the PET modality preprocessing methods \citep{a301,a302}.

\textbf{D. Other Modalities Preprocessing}

fNIRS and ECoG are two other modalities for functional neuroimaging of the brain employed by researchers for epileptic seizures detection \citep{a103,a104,a105}. Essential preprocessing steps in these modalities are similar to those of EEG signals and include noise reduction, normalization, and windowing of signals.

\subsubsection{Review of Deep Learning Techniques}

In recent years, with the increased availability of large datasets, methodologies rooted in DL techniques are poised for making a significant improvement in the diagnosis of various neurological disorders, including epileptic seizures. The DL-based CAD systems enable physicians to make better-informed decisions based on the recorded patient neuroimaging modalities. Figure \ref{fig:5} illustrates different types of DL architecture. It shows that CNNs \citep{a60}, GANs \citep{a64}, RNNs \citep{a60}, AEs \citep{a60}, DBNs \citep{a63}, CNN-AE \citep{a66}, and CNN-RNN \citep{a67} are the main DL architectures used for epileptic seizures detection. Among those, 2D-CNN and 1D-CNN are the most widely used DL architecture in the field of epileptic seizures (Tables \ref{tab:related} and \ref{tab:rel2}). This is due to the impressive achievements of CNNs architectures in other fields, including biomedical signal processing and medical imaging. In the rest of this section, we review the major DL network architectures and their variants. 

\textbf{A. 1D and 2D-CNNs}

\begin{figure}[h]
    \centering
    \includegraphics[width=\textwidth ]{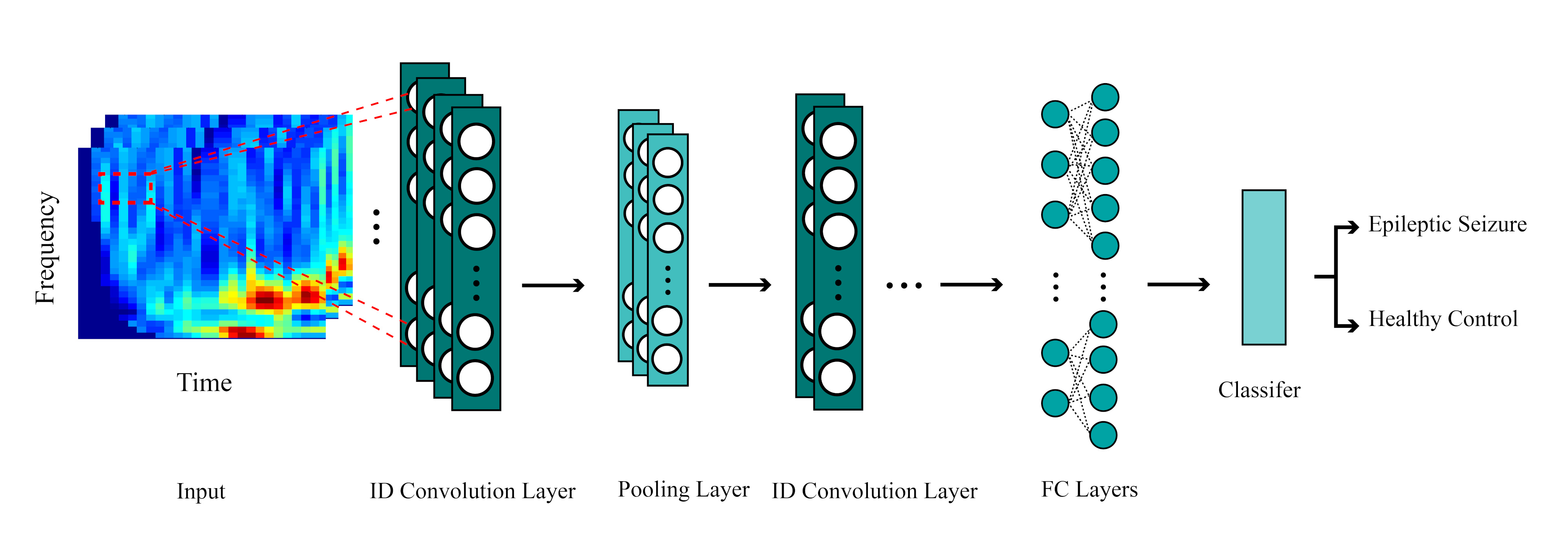}
    
    \caption{A typical 1D-CNN for epileptic seizure detection.}
    \label{fig:7}
\end{figure}

The idea of using neural net like algorithms has been around for decades, yet many limitations have stopped them from being useful in machine learning. With the famous AlexNet paper \citep{a60}, neural nets have resurfaced once again in the past decade. Adding some knowledge to the network structure, i.e., the fact that patterns are presented in spatial localities, led to convolutional layers, and by fixing the convolutional filters, the decrement in parameters made it possible for networks to train properly \citep{a60,n3,n4}. 2D-CNNs have been widely used since their first introduction, and their variant, 1D-CNNs, have also been applied vastly for signal processing tasks \citep{a106,a107}. Figure \ref{fig:7} shows a general form of a 2D-CNN used for epileptic seizure detection.

\textbf{B. Generative Adversarial Networks (GANs)}

\begin{figure}[h]
    \centering
    \includegraphics[width=\textwidth ]{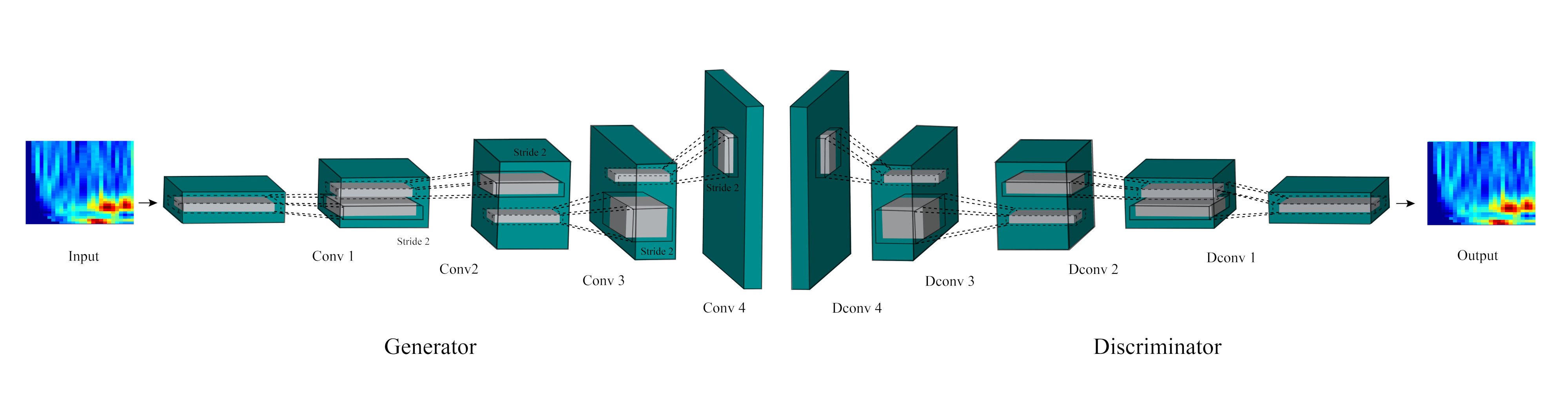}
    
    \caption{A typical GAN architecture for epileptic seizure detection.}
    \label{fig:8}
\end{figure}

In 2014, Goodfellow et al. \citep{a60} revolutionized the field of generative models by introducing Generative Adversarial Nets (GANs). The main contribution of GANs is their capability of generating high-quality images similar to the training dataset; GANs have been applied to signal \citep{a108,a109}, image \citep{a110,a111}, and many other data types in the past years \citep{n2}. Given the quality of generated data, GANs can be used for data augmentation and model pre-training \citep{a60}, helping to overcome one of the main issues in biomedical machine learning, the limited size of datasets. The general GAN architecture is shown in Figure \ref{fig:8}.

\textbf{C. Pre-Train Networks}

\begin{figure}[h]
    \centering
    \includegraphics[width=\textwidth ]{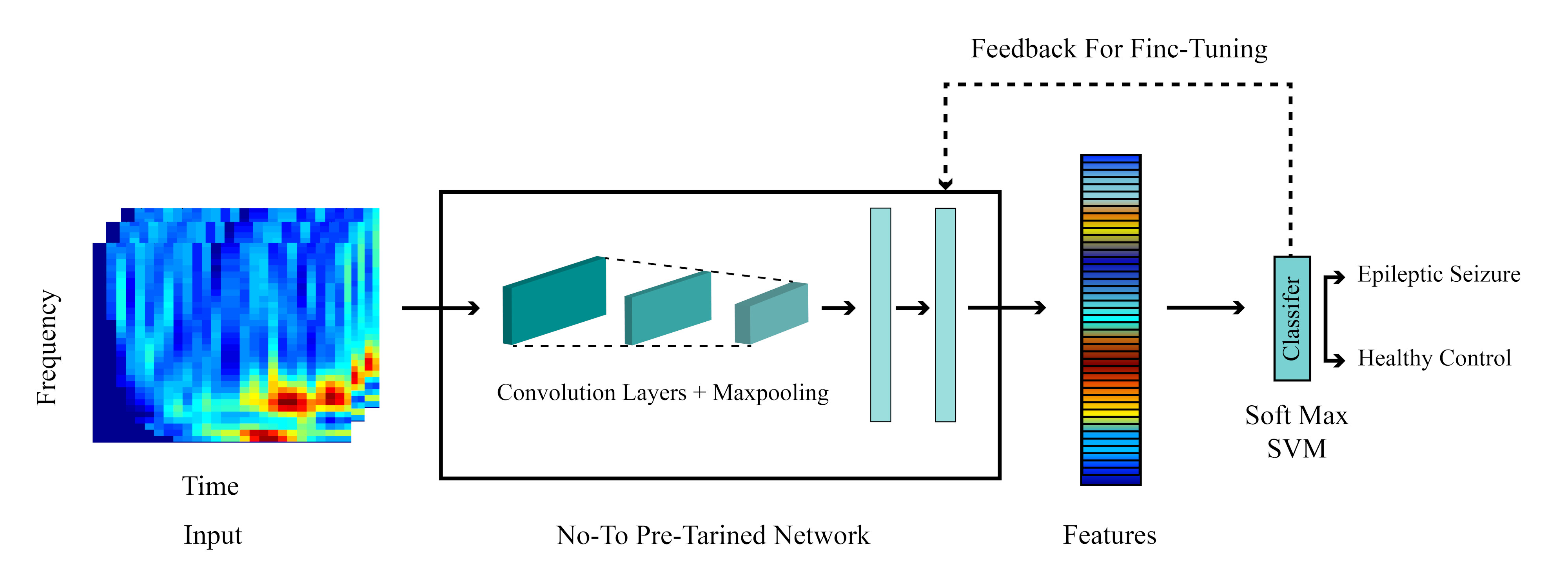}
    
    \caption{A typical deep pre-train network for epileptic seizure detection.}
    \label{fig:9}
\end{figure}

Deep neural nets usually have a tremendous amount of parameters; thus, they require enormous datasets for proper training. This is generally challenging in biomedical data processing due to small dataset sizes. However, one method used commonly to overcome this issue is to fine-tune previously trained networks. In this method, first, a DNN is trained on a big dataset, such as ImageNet, then the last layer, classifier, is removed. After that, as illustrated in figure \ref{fig:9}, its weights are fine-tuned using the primary dataset, or it is merely used as a feature extractor \citep{a60}. 

\textsc{AlexNet}

As the first famous DL network, AlexNet is still the center of attention in many studies \citep{a112}. In this network, two new perspectives dropout, and local response normalization (LRN), are used to help the network learn better. Dropout is applied in two FC layers employed in the end. On the other hand, LRN, utilized in convolutional layers, can be employed in two different ways: Firstly, applying single channel or feature maps, where the same feature map normalizes depending on the neighborhood values and selects the N×N patch. Secondly, LRN can be exploited across the channels or feature maps \citep{a60,a112}. 

\textsc{VGG}

Created by Visual Geometry Group in 2014, VGG is one of the pioneers of deep neural net structures; however, this famous structure is still extensively used and popular among researchers \citep{a113}. Many argue that this is due to its straight forward design and also ease of applying this network for transfer learning \citep{a113}. Two variants of VGG are mostly used for transfer learning, namely, VGG-16 and VGG-19 (number stands for the number of layers); also, they are applied in various fields, ranging from face recognition \citep{a114} to brain tumor classification \citep{a115}.
\textsc{GoogleNet}

Different receptive fields, generated by various kernel sizes, form layers called "Inception layers," which are the building block of these networks. Operations generated by these receptive fields find correlation patterns in the novel feature map stack \citep{a116}. In GoogLeNet, a stack of inception layers is used to enhance recognition accuracy. The difference between the final inception layer and the naïve inception layer is the inclusion of 1x1 convolution kernels, which performs a dimensionality reduction, consequently reducing the computational cost. Another idea in GoogleNet is the gradient injection, which aims to overcome the gradient vanishing problem. GoogLeNet comprises a total of 22 layers that is greater than any previous network. However, GoogLeNet uses much fewer parameters compared to its predecessors VGG or AlexNet \citep{a116,a60}.
\clearpage
\textsc{ResNet}

The idea behind ResNet was to overcome the issue of vanishing gradient by utilizing skip connections between blocks. This allowed the Residual nets to go deeper than regular networks; many varieties of these networks, with various sizes, such as 34, 50, and 152 have been created and applied in many tasks \citep{a117}. ResNet's main contribution was not the network or its state-of-the-art performance, but the network's building blocks, and similar blocks have been widely used in many other deep NN structures; as an example, Res2Net is an image segmentation network with a similar design to ResNet.

\textbf{D. 3D-CNN}

\begin{figure}[h]
    \centering
    \includegraphics[width=\textwidth ]{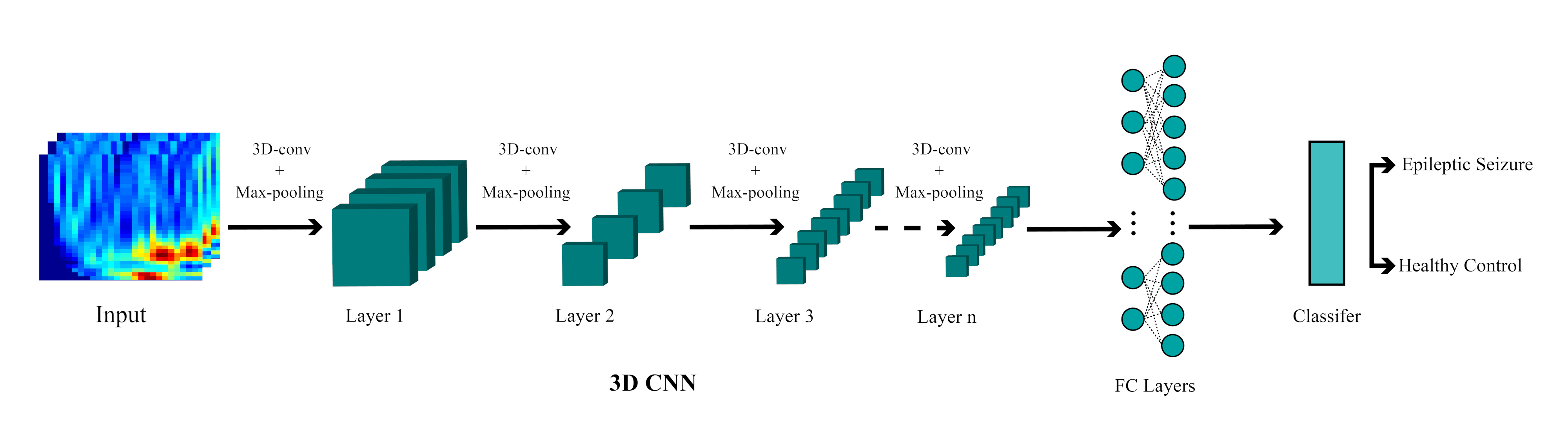}
    
    \caption{A typical 3D-CNN for epileptic seizure detection.}
    \label{fig:10}
\end{figure}

To overcome this, 3D-CNN was introduced. In 2D-CNN, many well-known structures such as VGG and GoogLeNet are available as a great starting point to construct the new structure upon them. However, creating 3D-CNNs can be challenging, considering there are not many famous 3D-CNN structures \citep{a118,a119}. Nevertheless, designed and trained properly, 3D-CNNs can find 3D patterns and achieve state-of-the-art performances. A typical 3D-CNN for epileptic seizure detection is shown in figure \ref{fig:10}. 

\textbf{E. Recurrent Neural Networks (RNNs)}

\begin{figure}[h]
    \centering
    \includegraphics[width=0.9\textwidth ]{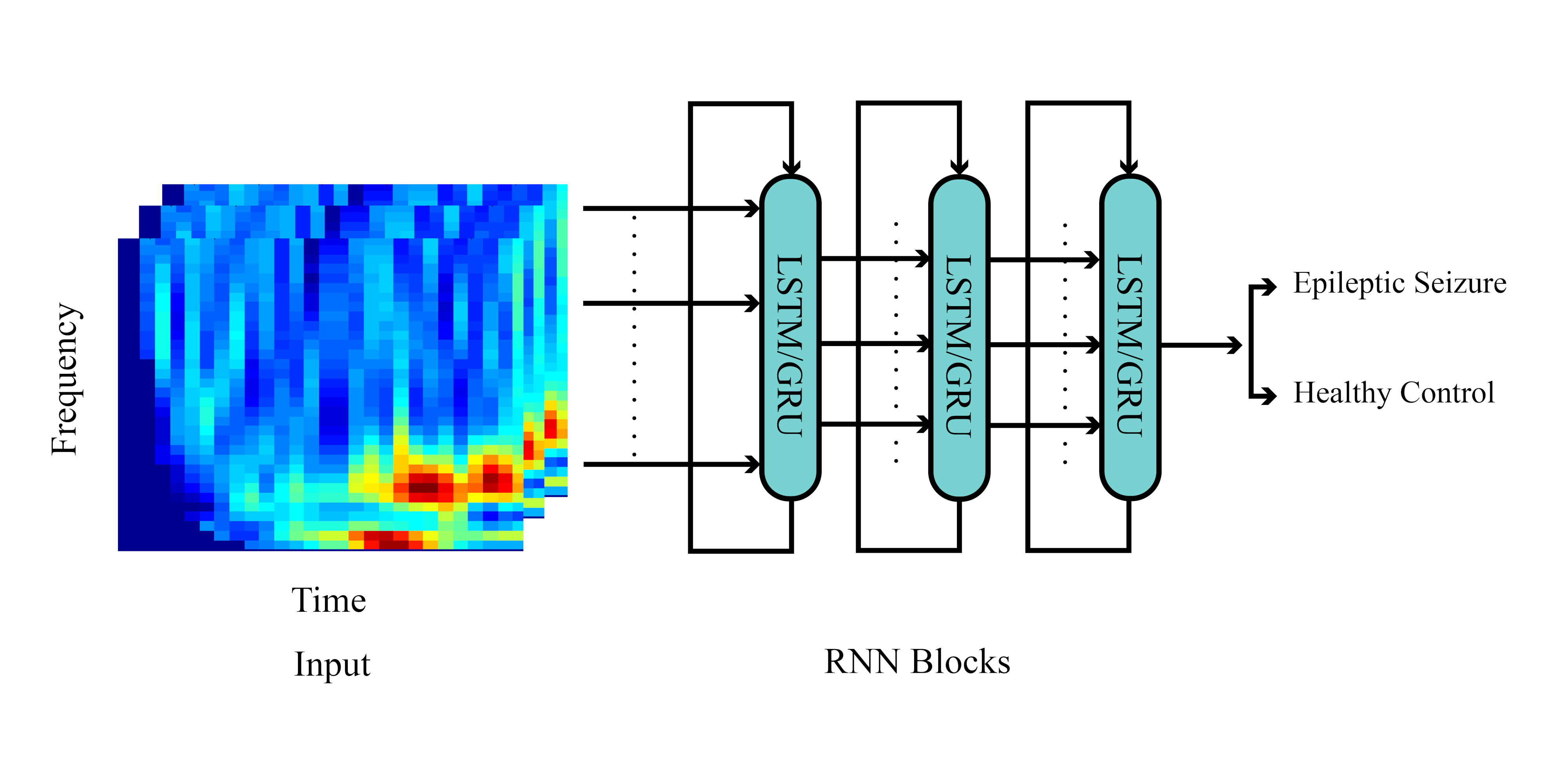}
    
    \caption{A typical RNN for epileptic seizure detection.}
    \label{fig:11}
\end{figure}

Many data forms, such as signal, have embedded patterns that cannot be characterized by local or spatial patterns. To create a model suitable for these datasets, researchers have created recurrent neural nets that, as stressed by the name, use the same group of neurons with a recurring scheme to process these data properly. Few variants of these networks, such as LSTM (long short term memory) and GRU, are created to find local and global patterns efficiently \citep{a120,a121}. The standard type of these networks is usually used as a baseline for creating models on signal processing and time-dependent datasets \citep{a122,a123}. However, a combination of these networks with convolutional layers is popular among researchers aiming to reach high performances with more complex models. A typical RNN for epileptic seizure detection is shown in figure \ref{fig:11}. 

\textbf{F. Deep Belief Networks (DBNs)}

Restricted Boltzmann Machine (RBM), the building block of Deep Boltzmann Machine (DBM), is an undirected graphical model \citep{a63}. The unrestricted Boltzmann machines are also similar; however, they may also have connections between the hidden units. DBNs are unsupervised probabilistic hybrid generative DL models comprising of latent and stochastic variables in multiple layers \citep{a63}. Moreover, a variation of DBN is called Convolutional DBN (CDBN), which is more suitable for images and signals, as it uses the spatial information of data \citep{a124}.

\textbf{G. Autoencoders}

\begin{figure}[h]
    \centering
    \includegraphics[width=\textwidth ]{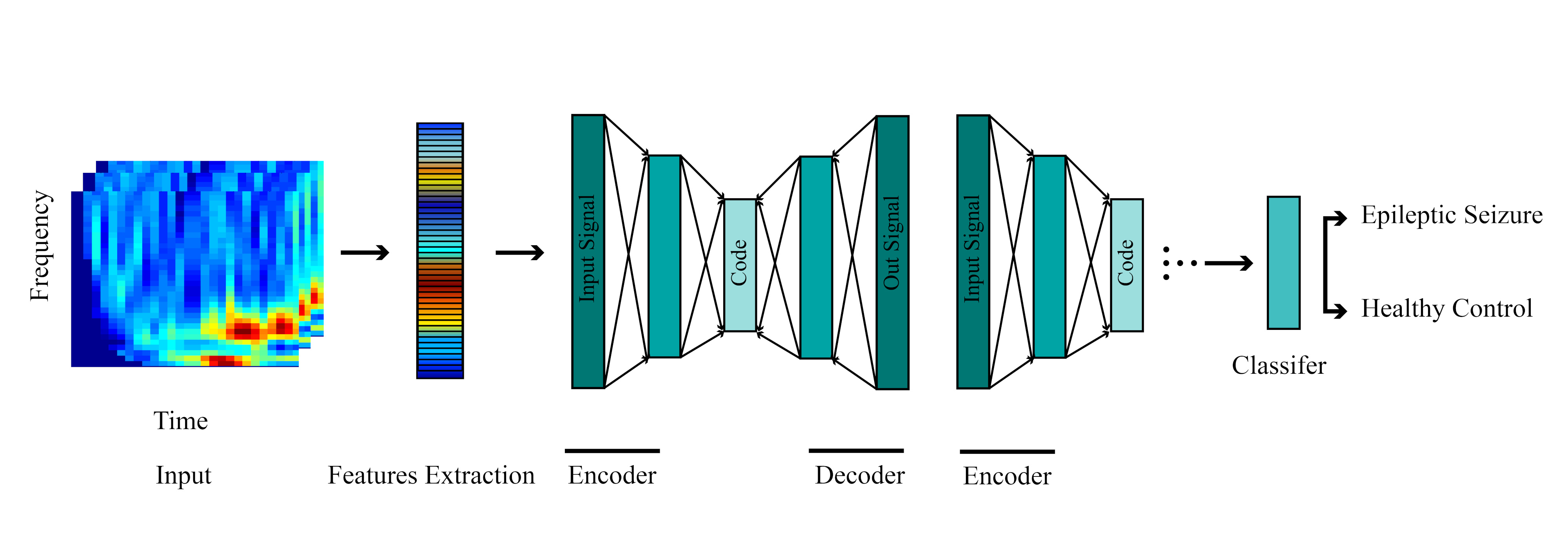}
    
    \caption{A typical AE for epileptic seizure detection.}
    \label{fig:12}
\end{figure}

AEs were one of the first groups of neural networks with practical use in machine learning \citep{a60}. Even with new advancements in DL, AEs have never lost researchers' attention and are widely used for dimensionality reduction and representation learning. The main idea behind AE is to map data to a smaller latent space and then back to the starting space with a minimum loss, thus reaching a mechanism to preserve essential aspects of data while reducing its dimensionality. Nowadays, many variations of AEs have been introduced with the goal of improving the base AE performance, such as stacked AE (SAE), denoising AE (DAE), and sparse AE (SpAE) \citep{a125,a126,a127}. A typical AE for epileptic seizure detection is shown in figure \ref{fig:12}. 

\textbf{H. CNN-RNN}

\begin{figure}[h]
    \centering
    \includegraphics[width=\textwidth ]{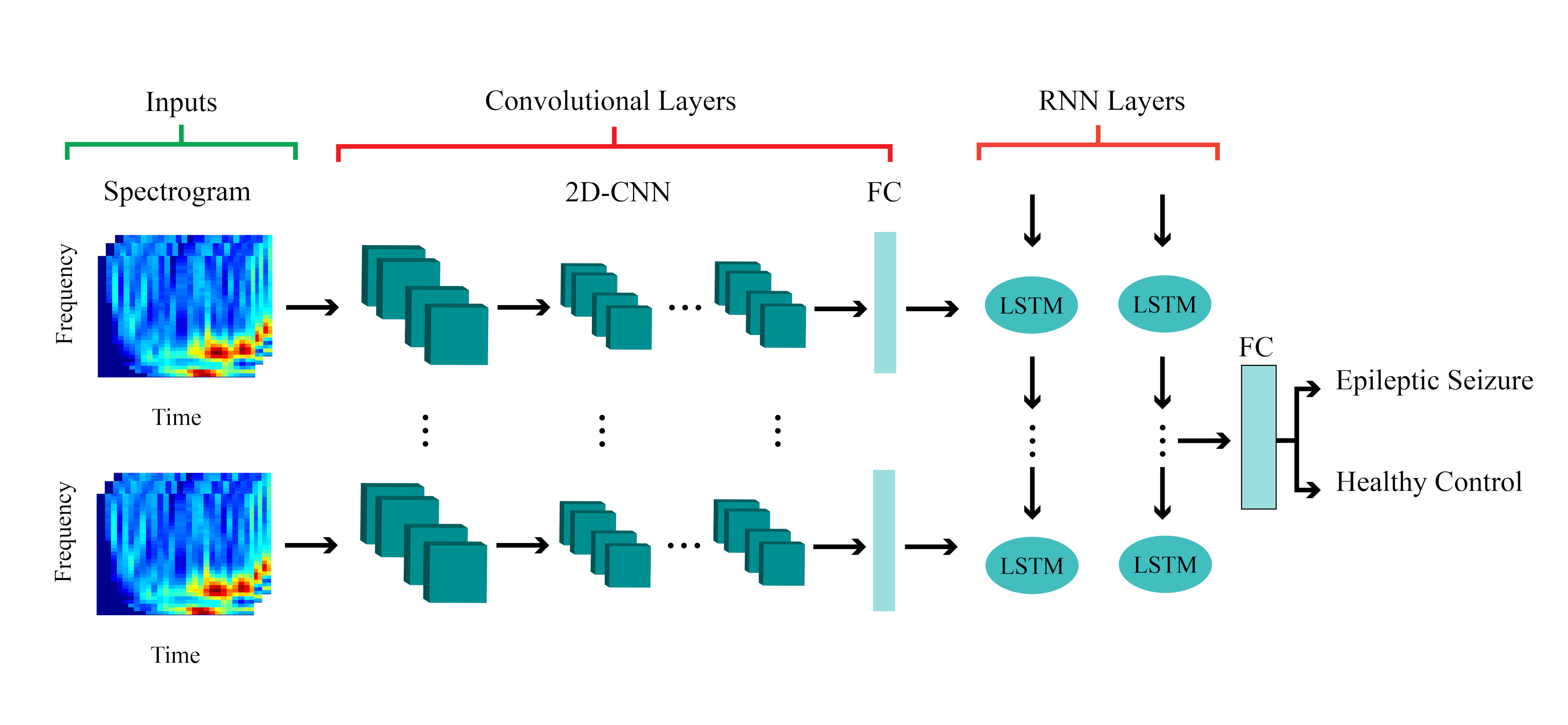}
    
    \caption{A typical CNN-RNN for epileptic seizure detection.}
    \label{fig:13}
\end{figure}

To combine RNNs and CNNs, researchers usually use convolutional layers in the first layers of the model to extract features and find local patterns, and they feed the output of these layers to RNNs to use their superiority for global pattern recognition \citep{a67}. The reasoning behind the noble performances of these models is a two-fold. First, convolutional layers empirically find local and spatial patterns considerably better than RNNs in signals. Second, adding convolutional layers allows RNN to see data with stride, hence finding more distanced patterns. By combining the output of convolutional layers and handcrafted features, CNN-RNNs are allowed to reach a state-of-the-art performance, in addition to learning a representation of data that overcomes handcrafted features' deficiencies \citep{a67}. A typical CNN-RNN for epileptic seizure detection is shown in figure \ref{fig:13}.  

\textbf{I. CNN-AE}

\begin{figure}[h]
    \centering
    \includegraphics[width=0.95\textwidth ]{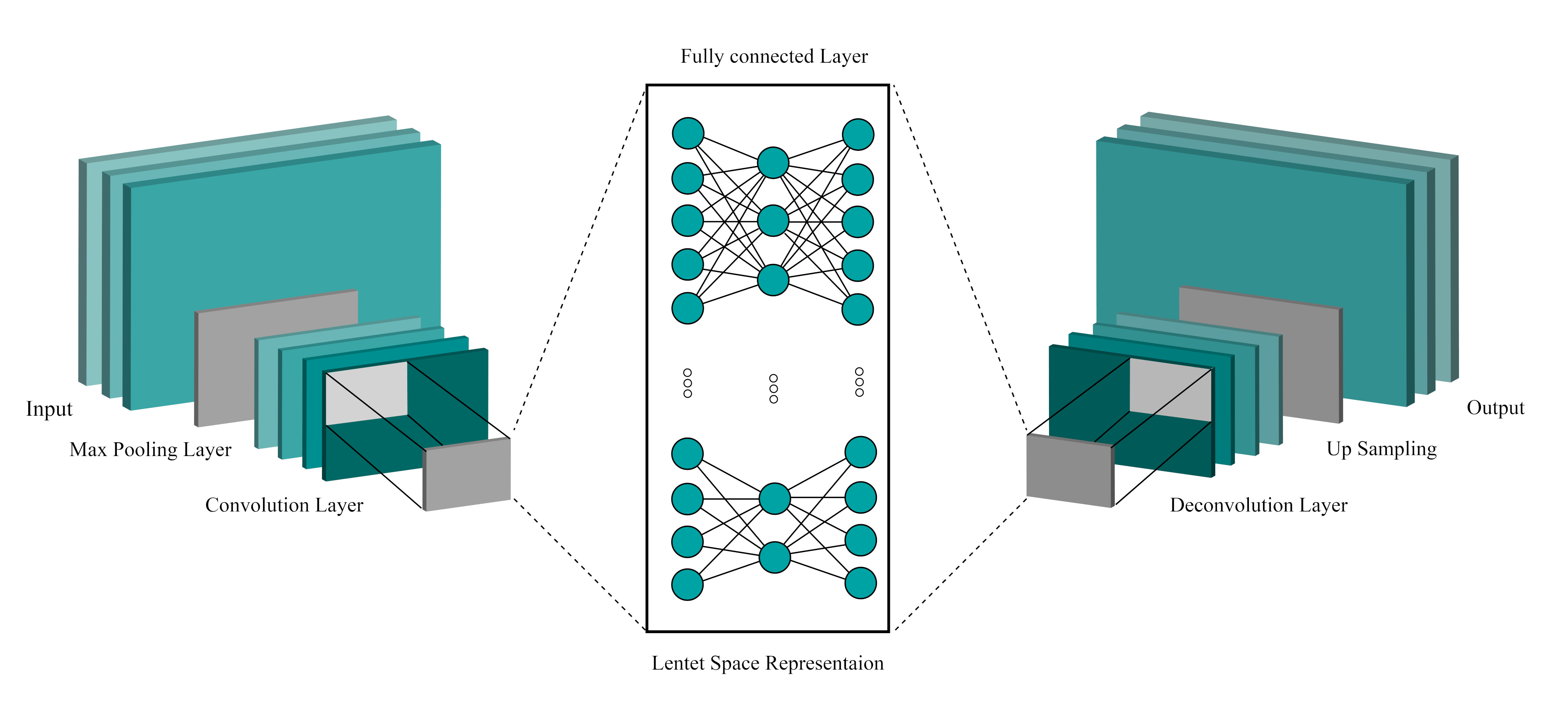}
    
    \caption{A typical CNN-AE for epileptic seizure detection.}
    \label{fig:14}
\end{figure}

Convolutional Autoencoder, or CNN-AE, is a DL based model that uses superiorities of convolutional layers to learn a representation of input unsupervised \citep{a66}. Base AEs are not suitable for raw representation learning, i.e., learn a representation of data without any added knowledge. This is due to the large number of learnable parameters that stops the network from learning anything useful. However, using convolutional layers, parameters are reduced dramatically, and networks can be appropriately trained \citep{a66}. A combination of this model with other ones, such as DAE, can lead to complex models with state-of-the-art performances \citep{a128,a129}. A typical CNN-AE for epileptic seizure detection is shown in figure \ref{fig:14}. 

\subsubsection{Other Neuroimaging Modalities for Epileptic Seizure Detection}

\textbf{A. Medical Imaging}

In the medical imaging literature, many researchers have focused on the application of fMRI, sMRI, and PET modalities for epileptic seizure detection using DL methods. sMRI and fMRI neuroimaging modalities are more popular than PET among physicians and neurologists for detecting epileptic seizures \citep{a130,a131}. This has led to more research papers be conducted on fMRI and sMRI modalities for epileptic seizures detection. Therefore, we summarize the relevant works that leverage various PET and MRI-based modalities in table \ref{tab:rel2}. 

\textbf{B. EEG-fMRI}

Multimodal neuroimaging techniques give physicians very detailed information about the type of neurological disorders and their location on the brain. As such, it is essential to use these modalities to identify the central location (focus) of epilepsy in the brain. EEG-fMRI is one of the best multimodal techniques for epileptic seizures detection \citep{a132,a133}. The modality of EEG-fMRI along with the ResNet network was investigated for epileptic seizures detection in \citep{a297}. In the proposed ResNet architecture, the Softmax and Triplet functions are used for supervised classification, achieving 84.40\% specificity.

\textbf{C. EEG - fNIRS}

fNIRS uses infrared waves to monitor changes in blood oxygen levels in the brain, allowing imaging and analysis of active brain areas \citep{a134}. In this method, using special electrodes on the scalp, variations in oxy-hemoglobin (HbO) and deoxy-hemoglobin (HbR) are measured, which can be helpful in diagnosing a variety of brain diseases. In \citep{a144}, EEG-fNIRS combination modalities have been employed to detect epileptic seizures. The proposed LSTM-based based approach, with a Softmax classifier as the last layer, achieved 98.30\% accuracy in their case study.

\textbf{D. ECoG}

RaviPrakash et al. \citep{a135} introduced an algorithm based on DL for Electrocorticography based functional mapping (ECoG-FM) for eloquent language cortex identification. ECoG-FM's success rate is low compared to Electro-cortical Stimulation Mapping (ESM). The algorithm showed an improvement of 34\% over the existing ECoG-FM method with an accuracy of approximately 89\%. ECoG-FM method coupled with DL has the potential for state-of-the-art performances. This method can help the surgeons performing epilepsy surgery by removing the ESM hazards. Also, in part of the \citep{a298}, an ECoG modality has been considered for the detection of epileptic seizures. 2D-CNN and SVM were used for feature extraction and classification steps, respectively. 

\textbf{E. MEG}

MEG is a functional neuroimaging technique used to evaluate and analyze the structure of the brain to diagnose a variety of brain disorders. Due to its high operational costs, it is only used in exceptional cases. \citep{a336} Proposed a new technique, EMS-Net, for detecting epileptic spikes from MEG modality, with satisfactory results.

\section{Rehabilitation Systems Based DL Techniques}

In recent years, research in the field of design and construction of rehabilitation systems aimed at assisting people with a variety of neurological disorders has advanced significantly. Rehabilitation systems are of particular significance to assist patients. The major objective of these systems is to achieve real and accessible tools for different patients. These systems are important in two aspects: First, they continuously monitor the patient's condition and, in the occurrence of disease, perform some necessary work to improve the disease. In the second case, there is another category of these systems that constantly report the patient's vital signs to the specialist so that the patient is at lower risk of disease. In this section, various rehabilitation systems are presented to help patients with epileptic seizures. These tools include programs to diagnose epileptic seizures from non-medical modalities \citep{a313,a314}, Brain Computer Interface (BCI) systems \citep{a315}, Implantable \citep{a316}, and Cloud-Computing \citep{a317,a318,a319,a320} are discussed below.

\subsection{Non Neuroimaging Modality for Epileptic Seizure Detection}

In a study by Ahmedt et al. \citep{a313}, facial images have been used to diagnose epileptic seizures. To collect the dataset, epileptic patients were monitored for 2 to 7 days, and eventually, 16 patients with MTLE were randomly selected from the general data set by default. The DL architecture in this investigation is CNN-RNN and, the results reveal that they have achieved promising results.

\subsection{Brain Computer Interface}

BCI based on DL to detect epileptic seizures has been recommended in Hosseini et al.'s study \citep{a315}. In the proposed technique, SSAE and Softmax methods have been exploited to perform feature extraction and classification steps, respectively. In this study, they achieved 94\% accuracy.

\subsection{Implantable Based DL}

Kiral-Kornek et al. \citep{a316} proposed an online and wearable system in the body for epileptic seizures detection based on DL. The proposed system has low power consumption, long life, and high reliability. In the proposed approach system, the DL method is trained to distinguish pre-ictal signals from ictal and has a sensitivity of 69\%.

\subsection{Cloud Computing Based DL for Epileptic Seizures Detection}
\begin{figure}[h]
    \centering
    \includegraphics[width=0.95\textwidth ]{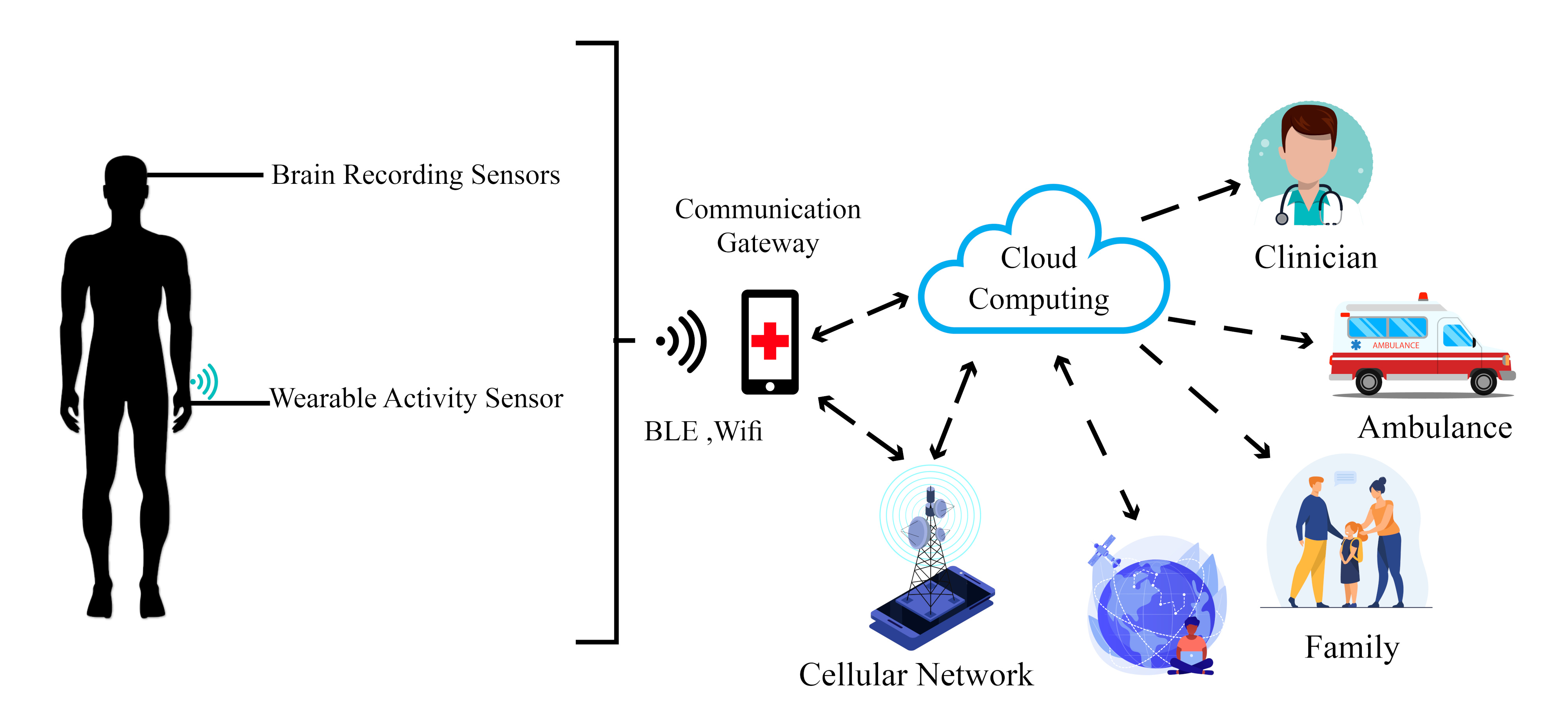}
    
    \caption{Cloud computing in helping patients with epileptic seizures.}
    \label{fig:15}
\end{figure}

With the advancement of information technology, performing heavy computational tasks at different times and places becomes a necessity. There is also a need for people to be able to easily fulfill their heavy computing tasks without owning expensive hardware and software. Cloud computing plays an important role in allowing users to process various data and store information outside of personal computers. The advantages of cloud computing have led to its accelerated application in various medical fields. An overview of cloud computing to help diagnose epileptic seizures is exhibited in Figure \ref{fig:15}. In the epileptic seizure detection field, research has been carried out using cloud computing, which we will describe below \citep{a317,a318,a319,a320,a321}.

\subsubsection{Cloud System Design Based DL and Smartphone}

Singh et al. \citep{a317} developed a commercial product for epileptic seizures detection, which involves user and cloud sections. The user section includes EEG headset, smartphone, WiFi system or, 4G network. The cloud segment also contains the dataset and the SAE algorithm for classifying EEG signals. EEG signals are recorded via a 14-channel Bluetooth headset and then transmitted to the patient's smartphone. Then, Android-based software transmits the recorded data to the cloud via WiFi or 4G internet connection. If the output of the classifier is pre-ictal, an alarm message containing the patient's geographical location is sent by the alarm system to the patient's telephone, family members' telephone, and the nearest hospital.

\subsubsection{Cloud System Based DL and Mobile Edge Computing}

Ali et al. \citep{a318} used the combination of DL with mobile edge computing to detect epileptic seizures. The objective of Edge Computing is to lessen the communication load of the cloud server and the edge device, which is specifically the main focus of this article. The proposed design assumes that the data has already been recorded and provided to the edge device, which is mobile. Next, receiving the raw data by mobile phone, they are partially processed and then sent to the cloud. Other processes continue to epileptic seizures detection in the cloud, and then the result is sent to the mobile phone.

\subsubsection{IoT Based Healthcare}

An IoT-based healthcare framework and DL to help patients with epileptic seizures are introduced in \citep{a319,a320}. In \citep{a319}, the function of two adopted cloud systems is employed, one of which sends EEG signals, and the other sends other vital information such as movements and emotions. With the cognitive module, the patient's vital signs are supervised online and then fed to the CNN network input. Finally, patient status and EEG signal analysis results are shared with medical providers. In the recommended approach, emergency help is provided if the patient is in critical condition.

\subsubsection{Mobile Multimedia Framework}

In a study by Muhammad et al. \citep{a321}, a technique based on mobile multimedia healthcare was proposed to help patients with epileptic seizures. In the proposed method, DL and the CHB-MIT dataset are utilized to detect epileptic seizures. Finally, the algorithms adopted are implemented on a module. Experimental results show the achievement of 99.02\% accuracy and 92.35\% sensitivity parameters.

\section{Discussion}
Today, many people worldwide suffer from epileptic seizures, and their daily activities are faced with serious challenges. So far, numerous clinical and screening procedures have been proposed to treat and diagnose epileptic seizures. Among the screening methods, EEG, fMRI, sMRI, and PET modalities are more important for epileptic seizures detection for physicians than other techniques. Applying DL techniques and neuroimaging modalities are crucially significant in epileptic seizure detection. In this paper, conducted researches on the diagnosis of epileptic seizures using DL methods are studied. Also, in the papers reviewed, practical applications in this field have been mentioned.

Diagnosis of epileptic seizures based on EEG modalities as well as medical imaging techniques are summarized in Tables \ref{tab:related} and \ref{tab:rel2}. These tables provide each research information, including dataset, modality, preprocessing techniques, DL network input, DL network, classification algorithm, K-Fold evaluation, and finally, various evaluation parameters.

\begin{figure}[h]
    \centering
    \includegraphics[width=0.57\textwidth ]{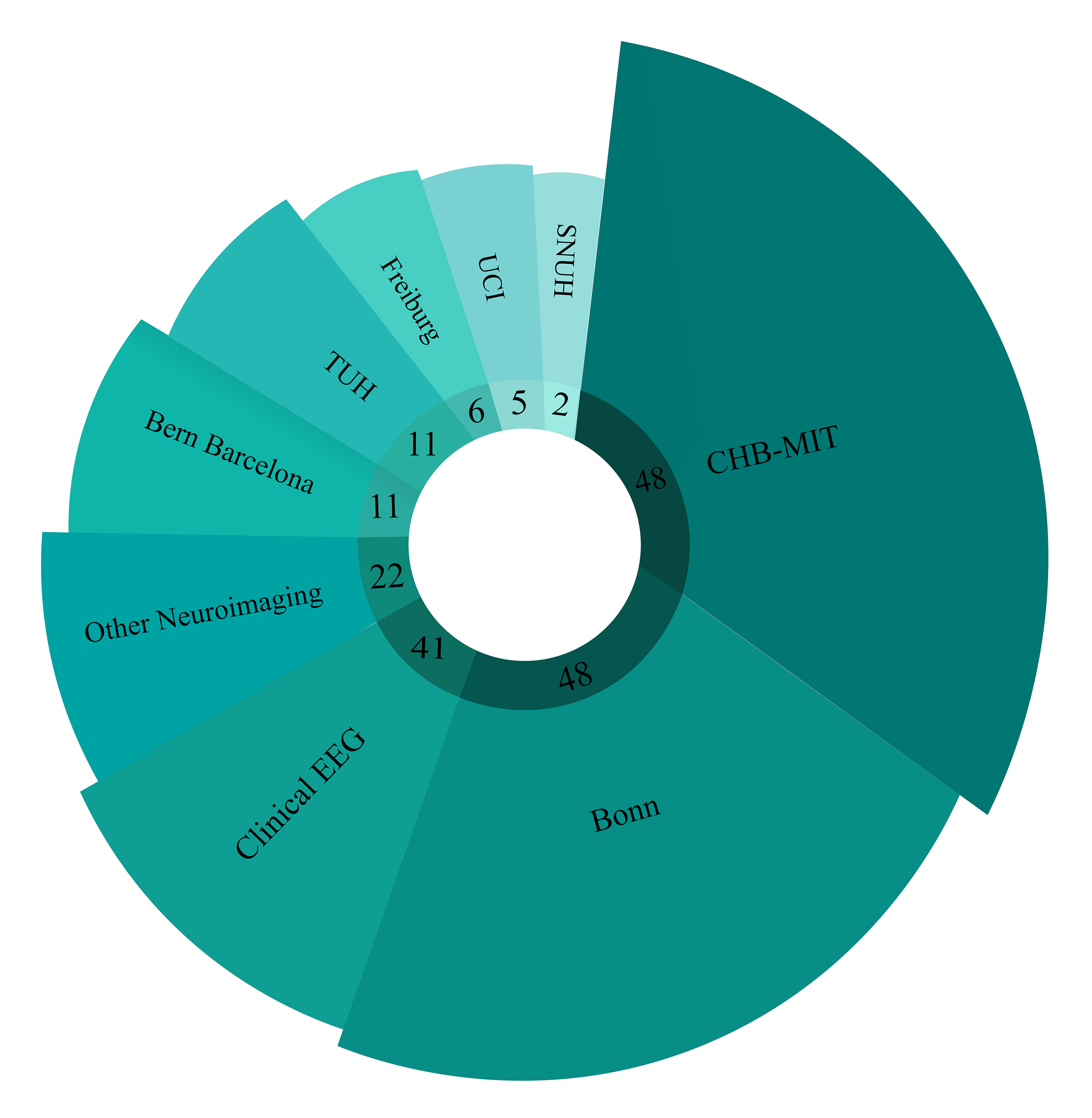}
    
    \caption{Number of studies published for epileptic seizures detection using different datasets.}
    \label{fig:16}
\end{figure}

The most important datasets available used for diagnosing epileptic seizures are provided in Table \ref{tab:data}. It is observable that the majority of them take advantage of EEG modalities. The total number of datasets utilized in epileptic seizure investigations is shown in Figure \ref{fig:16}. As can be observed, the Bonn dataset is the most widely used by researchers. This is because this database is preprocessed and can be easily employed for research.

\begin{figure}[h]
    \centering
    \includegraphics[width=0.57\textwidth ]{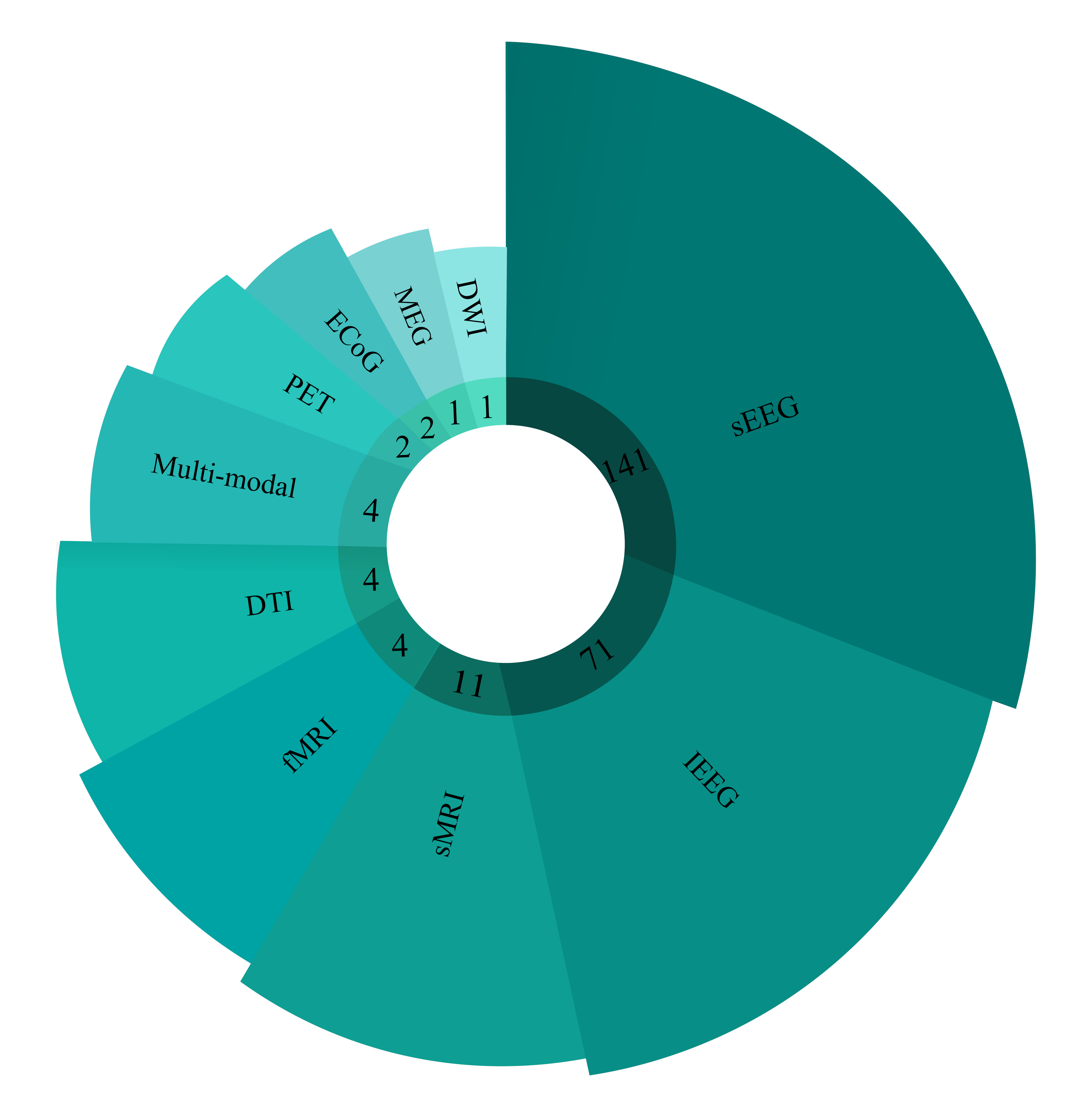}
    
    \caption{Number of studies published for epileptic seizures detection using different neuroimaging modalities.}
    \label{fig:17}
\end{figure}

Different neuroimaging modalities are applied to diagnose epileptic seizures. Detailed information on the types of neural modalities for diagnosing epileptic seizures is given in the diagram \ref{fig:17}. According to diagram \ref{fig:17}, the sEEG modality has dedicated to itself the highest use in the research. This is due to the non-invasive nature of the sEEG modality, which exposes patients to fewer risks. Moreover, according to Figure \ref{fig:17}, IEEG modality is considered the second epileptic seizures detection approach.

\begin{figure}[h]
    \centering
    \includegraphics[width=0.57\textwidth ]{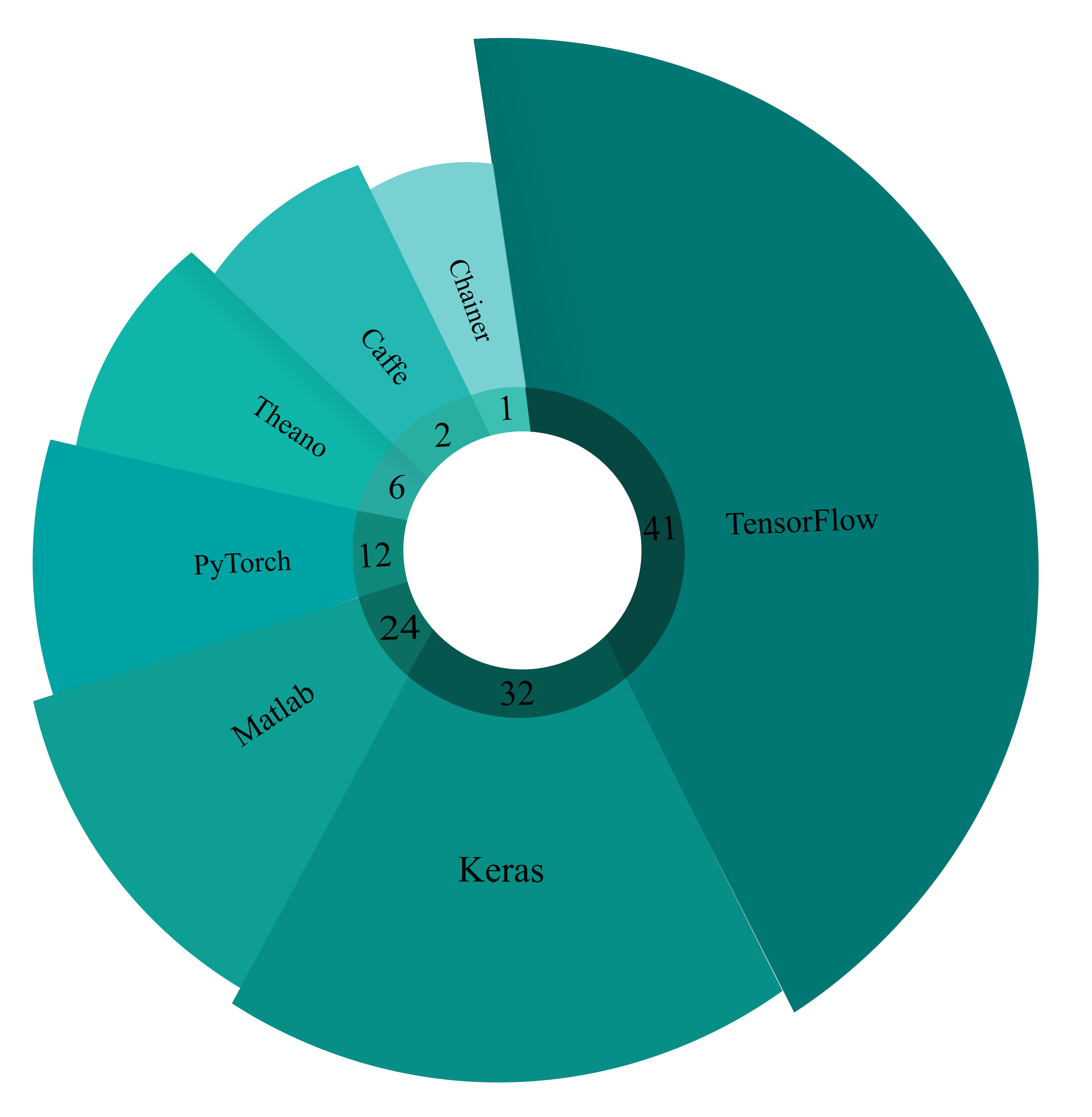}
    
    \caption{Number of DL tools used for epileptic seizures detection based on the published papers.}
    \label{fig:18}
\end{figure}

Numerous tools have been proposed to implement a variety of DL architectures, the main objective of which is to facilitate the simulation of these networks. Matlab, Keras, TensorFlow, PyTorch, Caffe, and Theano are the most well-known tools for the implementation of DL networks \citep{a332,a333}. The number of times each DL tool is used for epileptic seizure detection is illustrated in Figure \ref{fig:18}. The TensorFlow and Keras libraries are widely applied due to their continuous updating, high flexibility, and ease of use in implementing CADS epileptic seizures.

\begin{figure}[h]
    \centering
    \includegraphics[width=0.57\textwidth ]{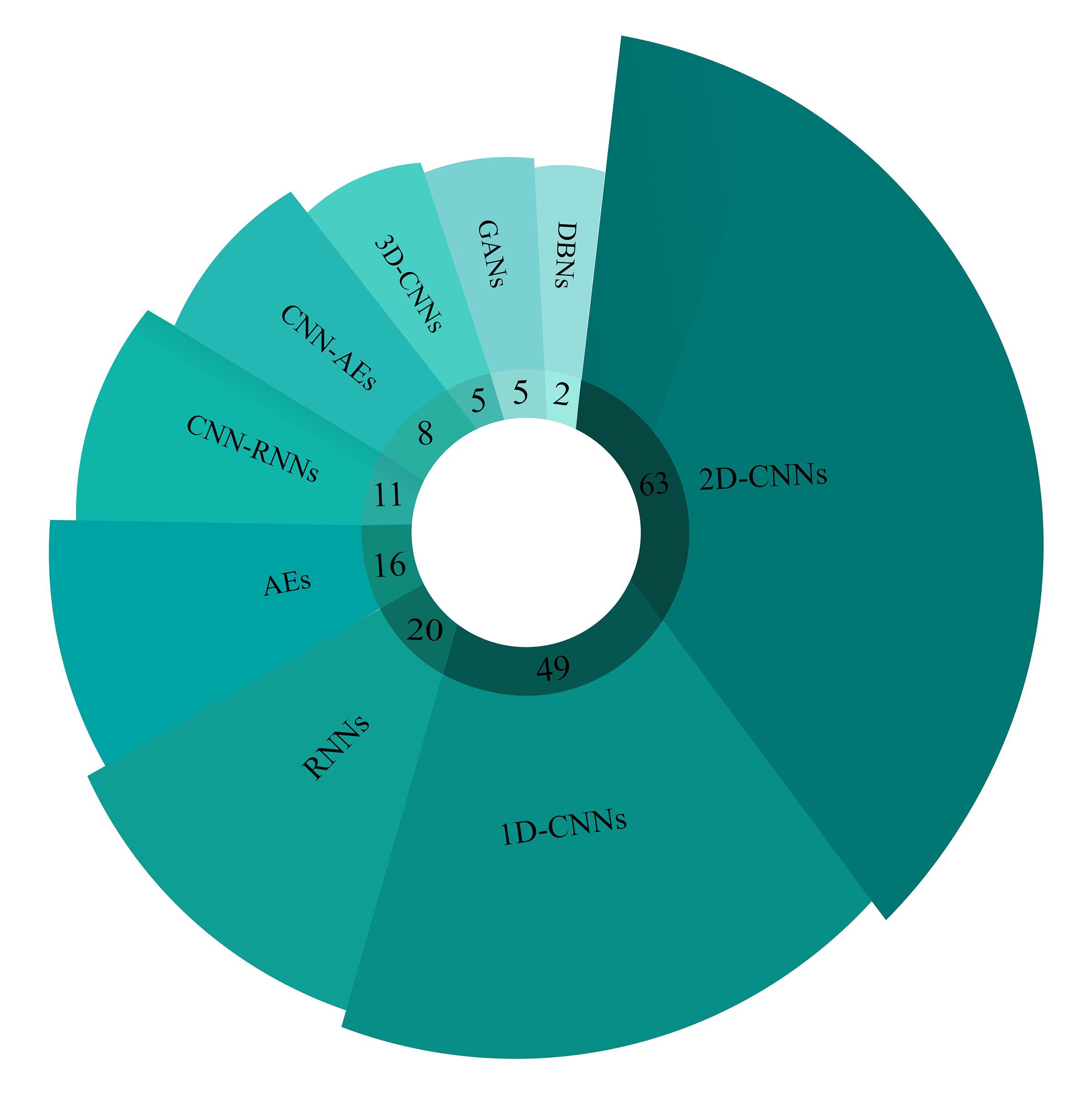}
    
    \caption{Number of DL architectures used for epileptic seizures detection based on published papers.}
    \label{fig:19}
\end{figure}

In tables \ref{tab:related} and \ref{tab:rel2}, the DL network types are discussed for epileptic seizures detection based on neuroimaging modalities. A variety of CNN models in various medical applications, especially the diagnosis of epileptic seizures, have reached promising results. Figure \ref{fig:19} display the total number of DL techniques for epileptic seizures detection.

Also, figure \ref{fig:19} shows the number of annual researches on the utilization of DL networks for epileptic seizures detection. Based on Figure \ref{fig:19}, the researchers have concentrated on different models of 2D-CNN and 1D-CNN. CNN architectures discover local spatial dependencies well; thus, these networks can be used to extract the necessary patterns from various modalities, including EEG signals. Furthermore, the patterns that CNNs learn are unchanged from relocation, and on the other hand, they can well train the hierarchy of feature space. In this article, not only the type of DL network in each research is discussed, but also in the table \ref{tab:app}, the implementation details of DL networks in each research are mentioned. 

\begin{figure}[h]
    \centering
    \includegraphics[width=0.57\textwidth ]{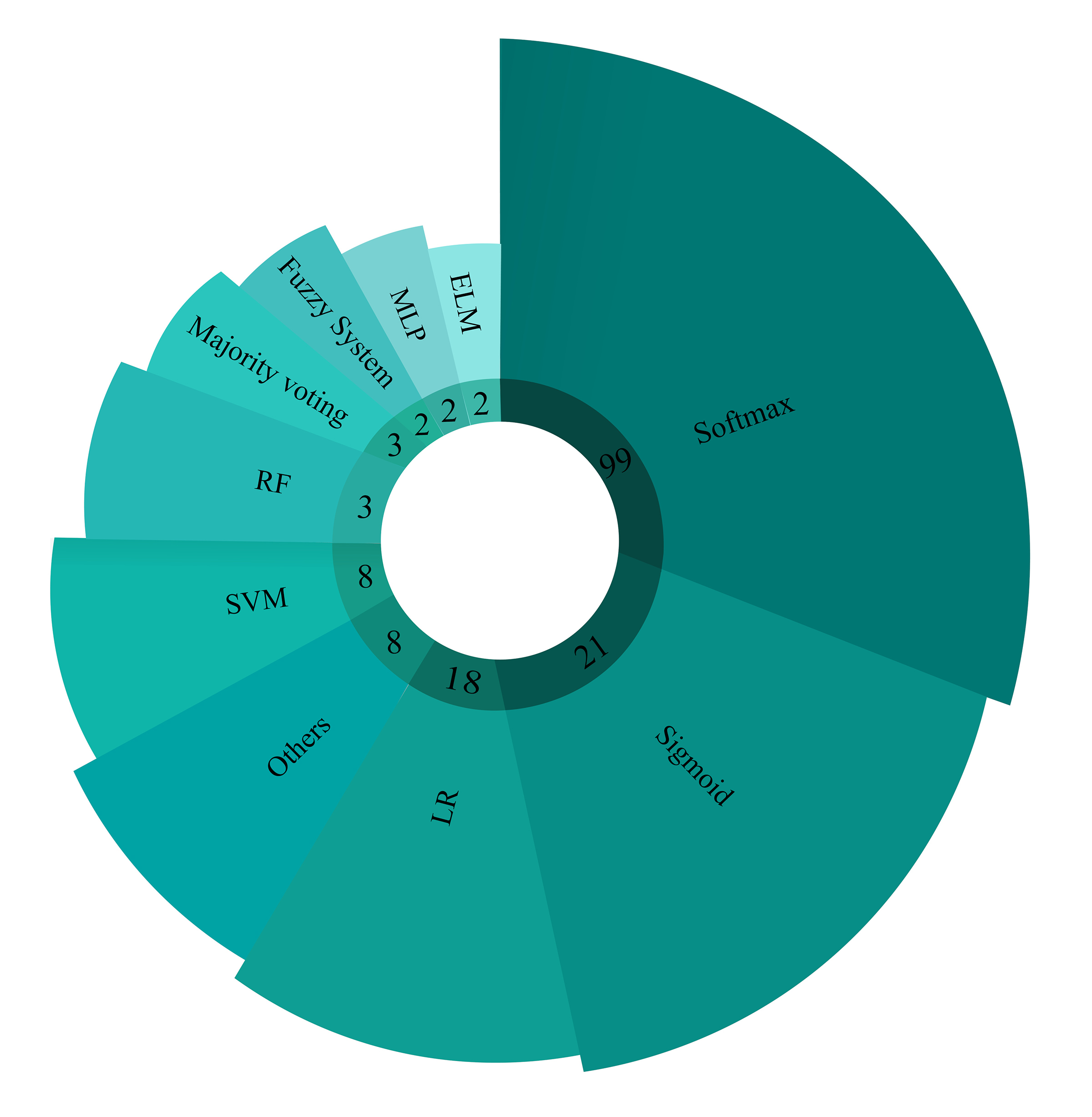}
    
    \caption{Illustration of the number of various classifier algorithms used in DL networks for automated detection of epileptic seizures.}
    \label{fig:20}
\end{figure}

Classification algorithms are the last part of the DL network. Figure \ref{fig:20} shows the number of classification algorithms used in DL networks based on Tables \ref{tab:related} and \ref{tab:rel2}. As can be seen, the Softmax algorithm \citep{a324} is the most popular in DL applications as a classification approach. Regarding the superiority of Softmax compared to other classifiers such as SVM \citep{a325}, we can remark its easy derivability, which makes it possible to apply it in the backpropagation algorithm. Also, compared to gradient descent methods, such as exploiting Sigmoid for classification purposes, Softmax provides better performance due to the weights normalization between different classes.

\section{Challenges}

In this section, the challenges in epileptic seizures diagnosis using DL techniques have been described. The most important challenges concerning datasets, DL methods, and hardware resources are explained in detail below.

Available datasets for diagnosing epileptic seizures are mostly EEG type. However, available datasets from other functional and structural neural modalities have not been provided for investigations until now. For example, the fNIRS modality is one of the most inexpensive and most accurate procedures to diagnose a variety of neurological disorders \citep{a326,a327}. The lack of available fNIRS datasets for the diagnosis of epileptic seizures has given rise to confined research in this field. Additionally, sMRI and fMRI modalities are recognized as some of the most significant and accurate tools for diagnosing brain disorders \citep{a328,a329}. So far, no dataset on sMRI or fMRI modalities has been made freely available to researchers for epileptic seizures detection. Multimodality techniques such as EEG-fMRI or EEG-fNIRS have been investigated in the diagnosis of mental and neural disorders and, noteworthy results have been achieved \citep{a330,a331,a332,a333}. In order to diagnose epileptic seizures using multimodality approaches, insufficient research has been performed, the main reason being the deficiency of available and free datasets.

Available EEG datasets commonly follow two approaches to distinguishing seizures from normal or when they occur. However, there are different types of epileptic seizures, and diagnosing their type is a troublesome task for physicians. Therefore, contributing datasets with functional or structural neuronal modalities to diagnose different types of epileptic seizures is profoundly felt.

Regarding the utilization of DL models for epileptic seizures detection, several challenges must be examined before implementing these models for clinical applications. The first challenge of this category is the extensiveness and differences of seizure patterns in signals. This issue leads to collect very large datasets to make these models more robust to new patterns or a more feasible solution to apply few-shot learning techniques to improve these models' robustness. Another challenge is to investigate the transferability of the model implemented on various datasets. Various studies have succeeded in achieving very high accuracy on particular datasets, but before adopting these models in real-world applications, their performance requires to be evaluated with a different distribution of the training data. The final challenge in this scope is the lack of networks with a dedicated structure for diagnosing epileptic seizures performing as a benchmark. In the image processing field, networks such as VGG and AlexNet have served as benchmarks and, in addition to providing researchers with a highly effective evaluation tool, allowing them to easily track their work to evolve and improve on previous work, while most of the networks used in this field are modified derivatives of networks presented for ImageNet and not specifically designed to diagnose epileptic seizures.

The following challenges category refers to the presentation of rehabilitation systems in the diagnosis of epileptic seizures with the help of DL. Unfortunately, much concentration has not been carried out on designing rehabilitation systems like BCI. In some research papers, cloud computing technologies, IoT and, Healthcare have been studied to address the difficulties of patients with epileptic seizures using neuroimaging modalities. The most substantial research challenges in this field are the deficiency of multimodality datasets for these systems' better performance.

Furthermore, dedicated hardware design platforms for this research have not been yielded till then, which is another challenge. To date, most researchers have developed the hardware implementation of conventional machine learning algorithms to detect epileptic seizures timely. This issue has led to these hardware circuits can not be employed for epileptic seizures detection seriously. Hardware implementation of DL algorithms to diagnose epileptic seizures can give specialist physicians the hope that they can accurately and real-time diagnose epileptic seizures and their type. Hardware implementation of DL algorithms on field-programmable gate array (FPGA), application-specific integrated circuit (ASIC), etc. can address many difficulties and challenges for medical professionals. However, when designing hardware based on DL, a number of existing challenges can be addressed, such as reducing hardware resources, minimizing power consumption, and so on.

\section{Conclusion and Future Works}

Early detection of epileptic seizures is of particular importance to specialist physicians, and research in this field has grown significantly in recent years. In this paper, a comprehensive review of the diagnosis of epileptic seizures using neuroimaging modalities coupled with DL methods has been performed. In the discussion section, it was observed that sEEG datasets are most applied in epileptic seizures detection. In another section, it was perceived that different models of DL have been employed to diagnose this type of neurological disorder, among which CNNs have the highest number of studies. In most investigations, small datasets have often been used to diagnose epileptic seizures alongside pre-train deep networks. To improve the performance of these DL networks, it is better to provide a comprehensive dataset of medical signals. This enhances the performance of pre-train deep networks for diagnosing epileptic seizures. Increasing the efficiency and accuracy of CAD systems in epileptic seizures detection is of particular significance, but aforementioned, the data deficiency is a serious challenge. Another novel field for research is applying Zero-Shot learning techniques, which can result in promising results for the implementation of real epileptic seizure detection systems.

In another part of the paper, the types of hardware and applied programs for detecting epileptic seizures were presented. Cloud Computing, IoT, Healthcare, and wearable implants have recently been introduced coupled with DL techniques to aid people with epileptic seizures, and it is encouraging that more applied research will be conducted in the near future. Lack of adequate hardware resources is another reason hardening the practical implementation of these systems. For future work, it is expected that CADS based on DL will be implemented on a variety of dedicated hardware such as FPGA and ASIC for epileptic seizures detection. 

Responsive neurostimulation (RNS) and vagus nerve stimulation (VNS) diagnostic systems are a type of invasive implants that can be implanted in the human body and are programmed to detect and neutralize the onset of epileptic seizures \citep{a334,a335}. These systems still have open problems in accurately diagnosing epileptic seizures. Enhancing the accuracy of RNS and VNS based diagnosis and treatment systems based on DL techniques can be noteworthy as one of the future tasks.

Another procedure of diagnosis and prediction is from other vital human signals such as the heart. Designing and manufacturing invasive and non-invasive implants based on other vital signals of the human body along with DL methods is another recommendation for future work. 

In addition to other future directions, the use of more sophisticated methods in deep neural networks can itself be a path for future works. The use of deep metric \citep{nb1} methods to increase the informativeness of learned representations, few-shot learning \citep{nb2} methods and scalable networks \citep{nb3} for small dataset tasks, and newer data augmentation methods such as simple copy-paste \citep{nb4} can all be investigated.

\clearpage

\begin{landscape}
\pagestyle{empty}

\begin{center}
\tiny
\setlength\LTleft{-108pt}            
\setlength\LTright{0pt}           

\begin{longtable}{|c|c|c|c|c|c|c|c|c|c|c|c|c|c|c|}

\caption{\centering Summary of DL methods employed for automated detection of epileptic seizures.}
    \label{tab:related}\\
    \hline
    \multicolumn{1}{|c|}{\multirow{2}{*}{\shortstack[1]{\textbf{Work}}}} & \multicolumn{1}{c|}{\multirow{2}{*}{\shortstack[1]{\textbf{Dataset}}}} & \multicolumn{1}{c|}{\multirow{2}{*}{\shortstack[1]{\textbf{Modality}}}} & \multicolumn{2}{c|}{\textbf{Preprocessing}} & \multicolumn{1}{c|}{\multirow{2}{*}{\shortstack[1]{\textbf{Input Network}}}} & \multicolumn{1}{c|}{\multirow{2}{*}{\shortstack[1]{\textbf{Deep Tools}}}} & \multicolumn{1}{c|}{\multirow{2}{*}{\shortstack[1]{\textbf{Network}}}} & \multicolumn{1}{c|}{\multirow{2}{*}{\shortstack[1]{\textbf{K-fold}}}} & 
    \multicolumn{1}{c|}{\multirow{2}{*}{\shortstack[1]{\textbf{Classifier}}}} & \multicolumn{4}{c|}{\textbf{Performance Criteria (\%)}}\\
    \cline{4-5}
    \cline{11-14}
    \multicolumn{1}{|c|}{} & \multicolumn{1}{c|}{} & \multicolumn{1}{c|}{} & \multicolumn{1}{c|}{\textbf{Low Level}} & \multicolumn{1}{c|}{\textbf{High Level}} & \multicolumn{1}{c|}{} & \multicolumn{1}{c|}{} & \multicolumn{1}{c|}{} & \multicolumn{1}{c|}{} & \multicolumn{1}{c|}{} & \multicolumn{1}{c|}{\textbf{ACC}} & \multicolumn{1}{c|}{\textbf{Sens}} & \multicolumn{1}{c|}{\textbf{Spec}} & \multicolumn{1}{c|}{\textbf{F1-S}}\\
    \hline
\citep{a136} & Clinical & IEEG & Standard Preprocessing  & Spectrogram & 2D Spectrograms & NA & 2D-CNN & -- & LR & 87.51 & -- & -- & --\\
\hline
\citep{a137} & CHB-MIT & sEEG & Segmentation &STFT &2D Spectrograms &PyTorch &2D-CNN &-- &ELM&95.65 &95.85 &-- & --\\
\hline
    \end{longtable}
\clearpage
\setlength\LTleft{-102pt}            
\setlength\LTright{0pt}           
\begin{longtable}{|c|c|c|c|c|c|c|c|c|c|c|c|c|c|c|}

\caption{\centering Summary of related works done using medical imaging methods and DL.}
    \label{tab:rel2}\\
    \hline
    \multicolumn{1}{|c|}{\multirow{2}{*}{\shortstack[1]{\textbf{Work}}}} &  \multicolumn{1}{c|}{\multirow{2}{*}{\shortstack[1]{\textbf{Dataset}}}} &  \multicolumn{1}{c|}{\multirow{2}{*}{\shortstack[1]{\textbf{Modalities}}}} &  \multicolumn{1}{c|}{\multirow{2}{*}{\shortstack[1]{\textbf{Preprocessing}}}} &  \multicolumn{1}{c|}{\multirow{2}{*}{\shortstack[1]{\textbf{Preprocessing}\\ \textbf{Toolbox}}}} &  \multicolumn{1}{c|}{\multirow{2}{*}{\shortstack[1]{Input Network}}} &  \multicolumn{1}{c|}{\multirow{2}{*}{\shortstack[1]{DNN}}} &  \multicolumn{1}{c|}{\multirow{2}{*}{\shortstack[1]{\textbf{DNN}\\ \textbf{toolbox}}}} &  \multicolumn{1}{c|}{\multirow{2}{*}{\shortstack[1]{Classifier}}} &  \multicolumn{1}{c|}{\multirow{2}{*}{\shortstack[1]{K-fold}}} &  \multicolumn{4}{c|}{Performance criteria (\%)}\\
    \cline{11-14}
\multicolumn{1}{|c|}{} & \multicolumn{1}{c|}{} & \multicolumn{1}{c|}{} & \multicolumn{1}{c|}{} & \multicolumn{1}{c|}{} & \multicolumn{1}{c|}{} & \multicolumn{1}{c|}{} & \multicolumn{1}{c|}{} & \multicolumn{1}{c|}{} & \multicolumn{1}{c|}{} & \multicolumn{1}{c|}{\textbf{Acc}} & \multicolumn{1}{c|}{\textbf{Sens}} & \multicolumn{1}{c|}{\textbf{Spec}} & \multicolumn{1}{c|}{\textbf{F1-S}}\\
\hline

\end{longtable}

\setlength\LTleft{-60pt}            
\setlength\LTright{0pt}           
\begin{longtable}{|c|c|c|c|c|c|}

\caption{\centering Details for DL Networks for Epileptic Seizures Detection.}
    \label{tab:app}\\
    \hline
\textbf{Work} & \textbf{Networks} & \textbf{Network Details} & \textbf{Classifier} & \textbf{Optimizer} & \textbf{Loss Function} \\
\hline
\multicolumn{1}{|c|}{\multirow{2}{*}{\shortstack[1]{\citep{a136}}}} &  \multicolumn{1}{c|}{\multirow{2}{*}{\shortstack[1]{2D-CNN}}} &  \multicolumn{1}{c|}{\multirow{1}{*}{\shortstack[1]{1 Conv Layer + 1 FC Layer}}} &  \multicolumn{1}{c|}{\multirow{2}{*}{\shortstack[1]{LR}}} &  \multicolumn{1}{c|}{\multirow{2}{*}{\shortstack[1]{NA}}} &  \multicolumn{1}{c|}{\multirow{2}{*}{\shortstack[1]{NA}}} \\
\multicolumn{1}{|c|}{} & \multicolumn{1}{c|}{} & \multicolumn{1}{c|}{2 Conv Layers + 1 FC Layer} & \multicolumn{1}{c|}{} & \multicolumn{1}{c|}{} & \multicolumn{1}{c|}{} \\
\hline
\citep{a137} & 2D-CNN & \makecell{3 Conv Layers + 3 Max Pooling Layers + 3 BN Layers + 1 FC Layer + ReLU\\activation function} & ELM & NA & NA \\
\hline
\citep{a138} & \makecell{Combination of\\1D-CNN and\\2D-CNN} & 6 1D Conv Layers + 3 2D Conv Layers + 3 Max Pooling Layers + 2 FC Layers & Sigmoid & \makecell{Sum-Squared\\Error} & Proposed Loss Function \\
\hline
\citep{a139} & 2D-CNN & 6 Conv Layers + 6 Dropout Layers + 3 Max Pooling Layers + 2 FC Layers & Softmax & Adam & \makecell{Categorical Cross\\Entropy (CCE)} \\
\hline
\citep{a140} & SeizNet & \makecell{4 Conv Layers + 4 Max Pooling Layers + 5 Dropout Layers + 4 BN Layers + 1 FC\\Layer} & Dense Layer & Adam & \makecell{Binary Cross Entropy\\(BCE)} \\
\hline

\end{longtable}
\end{center}
\end{landscape}
\newpage

\bibliography{main}  
\end{document}